\pdfoutput=1

\documentclass[11pt]{article}

\usepackage[final]{acl}

\usepackage{times}
\usepackage{latexsym}

\usepackage[T1]{fontenc}

\usepackage[utf8]{inputenc}

\usepackage{microtype}
\usepackage{subfigure}
\usepackage{booktabs} 
\usepackage{amsmath}
\usepackage{tabularx} 
\usepackage{multirow}
\usepackage{hyperref}

\usepackage{inconsolata}

\usepackage{graphicx}

\usepackage{amsmath}
\usepackage{amssymb}
\usepackage{mathtools}
\usepackage{amsthm}

\usepackage{colortbl}

\usepackage[capitalize,noabbrev]{cleveref}

\usepackage{float}

\theoremstyle{plain}

\theoremstyle{definition}

\theoremstyle{remark}

\usepackage[textsize=tiny]{todonotes}

\newcommand{\bx}{\mathbf{x}}
\newcommand{\by}{\mathbf{y}}

\newcommand{\model}{\textsc{\textbf{DiffPO}}}

\title{\textsc{DiffPO}: Turbocharging Inference Time Alignment of Large Language Models with Diffusion-Style Preference Optimization
}

\title{\textsc{DiffPO}: Diffusion-styled Preference Optimization for Efficient Inference-Time Alignment of Large Language Models}

\author{
Ruizhe Chen~$^{1,2,3}$ \quad \quad
Wenhao Chai~$^{4}$ \quad \quad
Zhifei Yang~$^{5}$ \quad \quad
Xiaotian Zhang~$^{1}$ \\
\textbf{Joey Tianyi Zhou}~$^{6}$ \quad 
\textbf{Tony Quek}~$^{3}$ \quad \quad
\textbf{Soujanya Poria}~$^{6}$ \quad
\textbf{Zuozhu Liu}~$^{1,2}$ 
\\
\newline
\\
$^{1}$ Zhejiang Key Laboratory of Medical Imaging Artificial Intelligence \quad \\
$^{2}$ Zhejiang University \quad 
$^{3}$ SUTD \quad 
$^{4}$ Princeton University \quad 
$^{5}$ Peking University \quad \\
$^{6}$ Nanyang Technological University \quad
$^{7}$ A*STAR Centre for Frontier AI Research \quad 
}

\begin{document}
\maketitle
\begin{abstract}
Inference-time alignment provides an efficient alternative for aligning LLMs with humans. However, these approaches still face challenges, such as limited scalability due to policy-specific value functions and latency during the inference phase. In this paper, we propose a novel approach, Diffusion-styled Preference Optimization (\model), which provides an efficient and policy-agnostic solution for aligning LLMs with humans. By directly performing alignment at sentence level, \model~avoids the time latency associated with token-level generation.
Designed as a plug-and-play module, \model~can be seamlessly integrated with various base models to enhance their alignment. Extensive experiments on AlpacaEval 2, MT-bench, and HH-RLHF demonstrate that \model~achieves superior alignment performance across various settings, achieving a favorable trade-off between alignment quality and inference-time latency. Furthermore, \model~demonstrates model-agnostic scalability, significantly improving the performance of large models such as Llama-3-70B. Our model and code are available \href{https://github.com/zjuruizhechen/DiffPO}{here}.
\end{abstract}

\section{Introduction}

The alignment of large language models (LLMs) with human preferences has recently emerged as a focal area of research \cite{wang2023aligning, shen2023large}. Prominent techniques such as Reinforcement Learning from Human Feedback (RLHF) \cite{ouyang2022training} and Direct Preference Optimization (DPO) \cite{rafailov2024direct} have demonstrated substantial efficacy. However, these methods require the optimization of individual policies, posing challenges such as high consumption of training resources. 
Inference-time alignment \cite{mudgal2023controlled, han2024value} provides an efficient alternative through direct adjustment of the model's output distribution, thus avoiding the need for resource-intensive retraining. Despite its advantages, this approach still requires policy-specific value functions, limiting its scalability across different models. Additionally, the inference-time latency remains high, presenting further challenges to its practical deployment.

\begin{figure}[t] 
    \centering
    \includegraphics[width=0.48\textwidth]{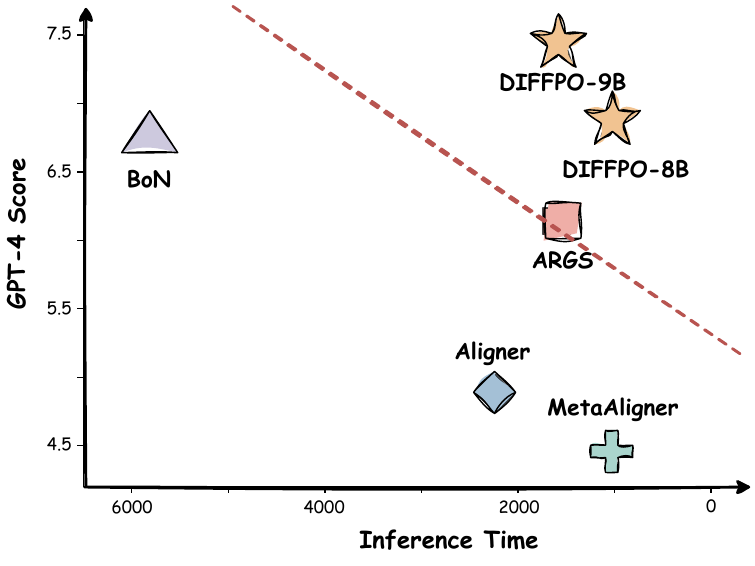}
    \caption{\textbf{Comparison with Inference-Time Methods.} Points closer to the \textbf{\textit{top-right}} indicate a superior trade-off between performance and inference time.}
    \label{fig: Trade_off}
    \vspace{-18pt}
\end{figure}

\begin{figure*}[h] 
    \centering
    \includegraphics[width=\textwidth]{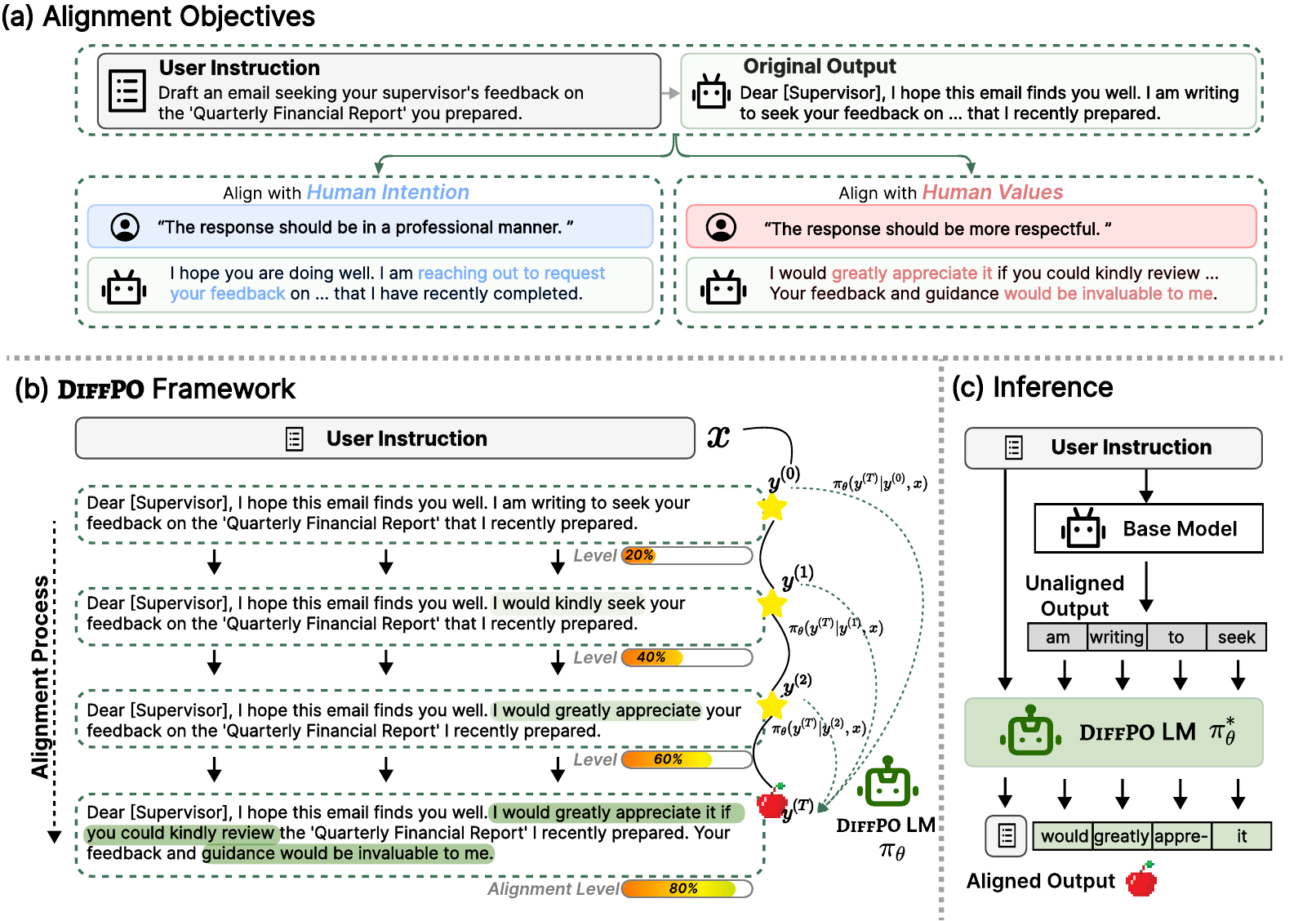}
    \caption{\textbf{Illustration of the \model~Framework.} (a) The objective of LLM alignment is to adjust the output of LLMs to reflect human values and intentions. In this process, preferences are considered at the \textbf{sentence level}, focusing on aspects such as the style and format of the complete output. (b) We propose Diffusion-style Preference Optimization (\model), which reconceptualizes the alignment process as a sentence-level denoising process, where the goal is to transform an unaligned sentence $\by^{(0)}$ into an aligned sentence $\by^{(T)}$ step by step.
    (c) Designed as a plug-and-play module, \model~can be directly integrated with the base model output and yield better alignment.}
    \label{fig: illustration}
    \vspace{-8pt}
\end{figure*}

In this paper, we investigate an efficient and policy-agnostic preference optimization method. We begin by reconsidering the objective of \textit{aligning} with humans~\cite{yao2023instructions, shen2023large}. As illustrated in Fig.~\ref{fig: illustration}(a), the alignment process operates at the sentence level, focusing on adjusting key components of the generated content, such as style or format, to better reflect human intentions or values. 
Inspired by the global controllability of the diffusion process~\cite{li2022diffusion, lyu2023fine}, we propose Diffusion-styled Preference Optimization (\model). 
\model~draws an analogy from the diffusion-based denoising process to model the iterative adjustment required for aligning human preferences, as shown in Fig.~\ref{fig: illustration}(b).
By employing parallel decoding~\cite{santilli2023accelerating, leviathan2023fast}, \model~directly predicts sentence-level transitions, thus avoiding the time latency associated with token-level generation. During the training phase, we optimize the \model~with an objective that maps generations with varying alignment levels to an aligned target, making it a policy-agnostic, plug-and-play module. 
The optimized \model~can then be seamlessly integrated with the output of the base model, enhancing its alignment level, as demonstrated in Fig.~\ref{fig: illustration}(c).

We evaluate the performance of \model~on several benchmark datasets, including AlpacaEval 2~\cite{dubois2024length}, MT-bench~\cite{zheng2023judging}, and HH-RLHF~\cite{bai2022training}. Empirical results demonstrate that \model~achieves superior alignment performance across various base models and settings. Compared to inference-time alignment techniques, \model~strikes an optimal trade-off between alignment performance and inference-time latency, as shown in Fig.~\ref{fig: Trade_off}. Additional experiments highlight the model-agnostic scalability of \model~across different base models. Specifically, \model-9B significantly enhances the performance of models such as Llama-3-70B and GPT-4o, showcasing its capability to improve weak-to-strong supervision.

\noindent The advantages of \model~can be summarized as:
\begin{itemize}
    \item \textbf{Model-agnostic.} 
    \model~is optimized to learn sentence-level refinement, independent of the specific base LLMs. This allows it to be applied across a variety of base LLMs. Furthermore, \model~does not require access to model parameters, which enhances its compatibility with API-based models and existing preference-aligned models.
    \vspace{-6pt}
    \item \textbf{Training and Inference Efficiency.} 
    As a post-inference alignment strategy, \model~adopts a one-for-all approach: it involves training \textit{one} single \model~and applying it \textit{for all} base models, thus significantly reducing the resource intensiveness associated with policy optimization. Moreover, by framing alignment as sentence-level prediction, \model~bypasses the time latency associated with token-level generation, thereby improving inference-time efficiency.
\end{itemize}

\section{Method}

\subsection{Preliminaries: Large Language Models}

\paragraph{Next-Token Prediction.}
The text generation of autoregressive large language models (LLMs) with prompt $\bx$ and response $\by$ can be modelled as a next-token prediction process. Given the input $\bx$, 
The language model $\pi(\cdot|\bx)$ autoregressively maps from current tokens $(\bx, \by_{1:n-1})$ to a distribution over the next token $\by_n$.
The maximum token, $N$, sets the length limit for LLM outputs, which conclude with an end-of-sentence (EoS) token $\by_{N}=$ EoS that ends the generation. The generated output $\by$ consists of predicted tokens $(\by_1, \by_2, . . . , \by_N)$.

\paragraph{Alignment of LLMs.}
During the alignment of LLMs, the objective is to optimize a language model $\pi_\theta$ that maximizes the user's preference~\cite{christiano2017deep, ouyang2022training,rafailov2024direct}:
\begin{align}
\label{RLHF_objective}
\max_{\pi_\theta} \mathbb{E}_{\substack{x \sim D, \by \sim \pi_\theta(\by|\bx)\\\by' \sim \pi_{\mathrm{ref}}(\by|\bx)} }& [ p(\by \succ \by' | \bx) \nonumber\\&- \beta D_{KL}(\pi_\theta \| \pi_{\mathrm{ref}}) ],
\end{align}
where \( p(\by \succ \by' | \bx) \) represents the preference, i.e., the probability that \( \by \) is preferred over \( \by' \) given the context \( \bx \), which can be generally represented by the reward function \( r \). The parameter \( \beta \) controls the deviation from the reference policy \( \pi_{\mathrm{ref}} \), which generally corresponds to the SFT model.



\paragraph{Parallel Decoding of LLMs.}

In comparison to next-token prediction, where token-level generation is performed sequentially to obtain a sentence, parallel decoding has demonstrated the capacity by enabling sentence-level generation and improving content quality~\cite{santilli2023accelerating, leviathan2023fast}. Concretely, supposing
\begin{equation}
f(\by_n, \by_{<n}, \bx) := \by_n - \arg \max_\by \pi(\by|\by_{<n}, \bx), \nonumber
\end{equation}
parallel decoding re-frames the LLM inference process as solving a system of nonlinear equations w.r.t. all tokens in a sentence $\by_n$ for $n = 1, \ldots, N$.
It can be solved in a parallel and iterative way:
\begin{equation}
\left\{
\begin{aligned}
\by_1^{(t+1)} &= \arg \max_\by \pi(\by \mid \bx) \\
\by_2^{(t+1)} &= \arg \max_\by \pi(\by \mid \by_1^{(t)}, \bx) \\
& \vdots \\
\by_N^{(t+1)} &= \arg \max_\by \pi(\by \mid \by_{<N}^{(t)}, \bx)
\end{aligned}
\right.
\end{equation}
In this way, for one forward pass of the LLM at time $t$, we can obtain the next sentence $\by^{(t+1)}$ based on the previous one $\by^{(t)}$.

\subsection{Diffusion-styled Preference Optimization}

\paragraph{Motivation.}
The goal of LLM alignment is to align the outputs of LLMs with human values or intentions~\cite{yao2023instructions}. In this process, preferences are defined at the \textbf{sentence-level}, focusing on the style or format of complete generated answers, as illustrated in Fig.~\ref{fig: illustration}(a). However, the generation of these responses occurs at the token level, following the next-token prediction pattern inherent in LLM modeling. This requires existing alignment techniques to optimize preferences (or rewards) at the \textbf{token-level}, which complicates the learning process~\cite{andrychowicz2017hindsight, zhong2024dpo, zeng2024token}. This inconsistency prompts us to reconsider the formulation of the alignment process.



\paragraph{Reformulation.} 

Inspired by the potential benefits of the diffusion process in controllable text generation~\cite{gong2022diffuseq, han2022ssd, ye2024diffusion}, we draw an analogy between the aligning LLMs and the diffusion process. Specifically, we propose \textbf{Diffusion-styled Preference Optimization} (\model), which reconceptualizes alignment as a sentence-level denoising process. 
The denoising process $\pi$ gradually refines the initial unaligned output $\by^{(0)}$ by adjusting the format or style as a whole. This process ultimately produces the aligned output $\by^{(T)}$, as illustrated in Fig.~\ref{fig: illustration}(b). The sentence-level alignment process can be formulated as follows:
\begin{equation}
\pi(\by^{(0:T)}) := p(\by^{(0)}) \prod_{t=1}^T \pi(\by^{(t)} | \by^{(t-1)}, \bx),
\end{equation}
where $\by^{(0)}$ and $\by^{(T)}$ represent the initial unaligned and final aligned generations, respectively. The intermediate sequence $\by^{(1:T-1)}$ can be viewed as the unaligned generations progressively transitioning along the trajectory from $\by^{(0)}$ to $\by^{(T)}$. 

Assuming the existence of a reward model $r(\bx, \by)$, which captures how well the generated output $\by$ aligns with human preferences given the input $\bx$, the goal is to optimize a \model~model $\pi_\theta$. This model learns to take a sentence as input and predict the next sentence with a higher reward, as illustrated in Fig.~\ref{fig: illustration}(c). The goal can be expressed as follows:
\begin{align}
\pi_\theta(\by^{(t)} | \by^{(t-1)}, \bx) \propto p(\by^{(t-1)}, \bx){\text{exp}}(r(\bx, \by^{(t)})). \nonumber
\end{align}
By employing parallel decoding, the \model \ model directly performs sentence-level predictions.



\subsection{Consistency Optimization of \model}
\label{sec:Consistency Optimization}

Inspired by Consistency LLMs~\cite{kou2024cllms}, we propose to consistently map any intermediate (unaligned) generation \( \by^{(t)} \) to the aligned generation \( \by^{(T)} \). We jointly optimize the \model~model \( \pi_\theta \) with two losses: one aligns the intermediate generation with the aligned generation, and the other prevents the corruption of the autoregressive (AR) modeling in the base model, thereby maintaining the generation quality.

\paragraph{Consistency Loss.} For a prompt \( \bx \) with an unaligned generation $\by^{(t)}$, we directly guide the model to output $\by^{(T)}$ with $\by^{(t)}$ as the input by minimizing the following loss $L_{\mathrm{Con}}$=
\begin{align}
\label{gc_loss}
\mathbb{E}_{(\bx, \by^{(t)}, \by^{(T)}) \sim \mathcal{D}} \left[ \sum_{i=1}^N \mathrm{KL}(\pi_{\theta^-} (\by_{<i}^{(T)}, \bx) \| \pi_\theta (\by_{<i}^{(t)}, \bx)) \right]
\end{align} 
where \(\theta^- = \mathrm{stopgrad}(\theta)\) and $N$ denotes the length of generation. \( \mathrm{KL}(\cdot \| \cdot) \) denotes the forward KL distance between two distributions. 


\paragraph{AR Loss.} 
To prevent the corruption of the autoregressive (AR) modeling in the base model and maintain the generation quality, we incorporate the AR loss based on the generated sequence \( \by^{(T)} \):
\begin{align}
L_{\mathrm{AR}} = \mathbb{E}_{(\bx,\by^{(T)}) \sim \mathcal{D}} \left[ -\sum_{i=1}^N \log \pi_\theta(\by^{(T)}_i | \by^{(T)}_{<i}, \bx) \right].
\end{align}
The total loss with weight \( \omega \) is:
\begin{align}
\label{total_loss}
L(\theta) = L_{\mathrm{AR}} + \omega L_{\mathrm{Con}}.
\end{align}


\subsection{The Objective of \model~within RLHF}
In this section, we analyze the role of \model~in achieving the goal of RLHF. 
We start with the same RL objective as prior work, Eq.~\ref{RLHF_objective}, under a general reward function $r^*$. Following prior work~\cite{peng2019advantage, rafailov2024direct}, the optimal solution to the KL-constrained reward maximization objective in Eq.~\ref{RLHF_objective} takes the form:
$r^*(\bx, \by) = \beta \log \left( \frac{\pi^*(\by \mid \bx)}{\pi_{\mathrm{ref}}(\by \mid \bx)} \right) + \beta \log Z(\bx)$, 
where \( Z(\bx) = \sum_\by \pi_{\mathrm{ref}}(\by | \bx) \exp\left( \frac{1}{\beta} r^*(\bx, \by) \right) \) is the partition function. With Bradley-Terry model, we can represent the preference function as the difference of rewards for a preferred answer $\by_w$ and a dispreferred answer $\by_l$: 
\begin{align*}
    p&(\by_w \succ \by_l | \bx) = \sigma(r^*(\bx, \by_w) - r^*(\bx, \by_l)) \\
    &= \sigma \left(\beta  \log \frac{\pi^*(\by_w \mid \bx)}{\pi_{\mathrm{ref}}(\by_w \mid \bx)}  -\beta  \log \frac{\pi^*(\by_l \mid \bx)}{\pi_{\mathrm{ref}}(\by_l \mid \bx)} \right).
\end{align*}
Substitute by $\pi^*(\by \mid \bx)= \pi_{\mathrm{\model}}(\by \mid \by',x)$ $\pi_{\mathrm{ref}}(\by' \mid \bx)$, we obtain $p(\by_w \succ \by_l | \bx)$ equals to 
\begin{align}
\label{eq:parpo}
    \sigma \left( \beta \log \frac{\pi_{\mathrm{\model}}(\by_w \mid \by_l,\bx)}{\pi_{\mathrm{\model}}(\by_l \mid \by_l,\bx)}  - \beta \log \frac{\pi_{\mathrm{ref}}(\by_w \mid \bx)}{\pi_{\mathrm{ref}}(\by_l \mid \bx)} \right).
\end{align}
Note that the first term in Eq.~\ref{eq:parpo} is optimized through the consistency loss in Eq.~\ref{gc_loss} by maximizing the probability of predicting \( \by_w \). The second term depends only on \( \bx \), with \( \pi_{\mathrm{ref}} \) remaining constant. Moreover, the deviation from the base policy can be easily controlled, since \( \by_w \) is derived from \( \by_l \).

\textit{In summary, the objective of \model~as defined in Eq.~\ref{total_loss} aligns with the RLHF objective in Eq.~\ref{RLHF_objective}.} Furthermore, since \( \pi_{\text{\model}} \) is optimized independently from the base model \( \pi_{\text{ref}} \), it can be deployed in a model-agnostic manner.




\subsection{Practical Implementations}
\label{sec:Practical Implementations}

\paragraph{Generate Alignment Trajectories.}

To implement \model, we collect the alignment trajectory for each prompt, thereby forming an original training set $\mathcal{D}$. Specifically, for each prompt $\bx$ from the \textit{UltraFeedback} dataset~\cite{cui2023ultrafeedback}, we generate $T$ responses using different base models. We then employ \textit{ArmoRM}~\cite{ArmoRM} reward model to score these responses. The response with the highest score is selected as $\by^{(T)}$. The remaining five responses are ranked based on their scores to form $\by^{(0:T-1)}$. $T$ is set to 6.

\paragraph{Training and Inference.} 
During the training phase, we initialize our aligning model $\pi_\theta$ using three backbones of varying sizes: Gemma-2-it-2B/9B, and Llama-3-8B-Instruct. The \model \ model is optimized adhering to the optimization loss in Eq.~\ref{total_loss} with parameters $N=256$ and $w=10^3$. Given the variable lengths of generations in $\mathcal{D}$, we standardize their lengths through padding or truncation. In the inference phase, the optimized model $\pi_\theta^*$ is employed to align responses from the vanilla generations produced by base models. Appendix~\ref{sec: Appendix Experimental Setups} shows more implementation details.


\section{Experiment}
\begin{table*}[t]
\centering
\resizebox{.98\textwidth}{!}{
  \begin{tabular}{@{}l >{\centering\arraybackslash}m{1.3cm}
  >{\centering\arraybackslash}m{1.3cm}
  >{\centering\arraybackslash}m{1.3cm}
  >{\centering\arraybackslash}m{1.3cm}
  >{\centering\arraybackslash}m{1.3cm}
  >{\centering\arraybackslash}m{1.3cm}
  >{\centering\arraybackslash}m{1.3cm}
  >{\centering\arraybackslash}m{1.3cm}
  >{\centering\arraybackslash}m{1.3cm}
  >{\centering\arraybackslash}m{1.5cm}@{}}
    \toprule
    \multirow{4}{*}{\textbf{Method}} & \multicolumn{5}{c}{\textbf{Llama-3-SFT (8B)}} & \multicolumn{5}{c}{\textbf{Llama-3-Instruct (8B)}} \\
    \cmidrule(r){2-6} \cmidrule(lr){7-11}
    &\multicolumn{1}{c}{\textbf{MT-bench}} & \multicolumn{2}{c}{\textbf{AlpacaEval 2}} &\multicolumn{2}{c}{\textbf{HH-RLHF}} & \multicolumn{1}{c}{\textbf{MT-bench}} & \multicolumn{2}{c}{\textbf{AlpacaEval 2}} &\multicolumn{2}{c}{\textbf{HH-RLHF}}\\
    \cmidrule(r){2-2} \cmidrule(lr){3-4} \cmidrule(lr){5-6} \cmidrule(lr){7-7}\cmidrule(lr){8-9} \cmidrule(lr){10-11}
           & \textbf{GPT-4}  & \textbf{LC (\%)} & \textbf{WR~(\%)} & \textbf{Helpful} & \textbf{Harmless} & \textbf{GPT-4}  & \textbf{LC (\%)} & \textbf{WR~(\%)} & \textbf{Helpful} & \textbf{Harmless} \\
    \midrule
Base Model & 6.21 & 22.09 & 20.81 & 0.59 & 0.91 & 6.78 & 36.83 & 42.12 & 0.67 & 0.93 \\
\textit{w.} DPO & 6.59 & 29.84 & 36.77 & \underline{0.68} & 0.89 & 6.90 & 47.20 & 53.56 & \underline{0.74} & 0.92 \\
\textit{w.} SimPO & 6.62 & 32.27 & \underline{40.96} & 0.66 & 0.86 & \underline{7.05} & \underline{52.57} & \underline{58.33} & \textbf{0.75} & 0.92 \\
\textit{w.} BPO & 5.84 & 21.34 & 22.33 & 0.60 & 0.92 & 6.43 & 22.39 & 34.06 & 0.67 & 0.92 \\
\textit{w.} ARGS & 6.14 & 9.06 & 13.97 & 0.49 & 0.86 & 6.84 & 31.83 & 34.74 & 0.64 & 0.89 \\
\textit{w.} BoN & 6.79 & 35.14 & 32.26 & 0.62 & 0.92 & 6.89 & 45.10 & 49.94 & 0.67 & 0.92 \\
\textit{w.} Aligner & 4.88 & 20.41 & 17.15 & 0.60 & 0.91 & 4.82 & 32.53 & 32.69 & 0.67 & 0.96 \\
\textit{w.} MetaAligner & 4.46 & 19.81 & 18.23 & 0.52 & 0.89 & 4.50 & 20.75 & 19.08 & 0.52 & 0.91 \\
\arrayrulecolor{gray}
\midrule
\arrayrulecolor{black}
\textit{w.} \model-8B & \underline{6.96} & \underline{36.24} & \underline{40.96} & 0.62 & \underline{0.93} & 7.02 & 36.44 & 41.01 & 0.68 & \underline{0.93} \\
\textit{w.} \model-9B & \textbf{7.45} & \textbf{49.72} & \textbf{54.23} & \textbf{0.71} & \textbf{0.98} & \textbf{7.40} & \textbf{55.84} & \textbf{61.88} & 0.72 & \textbf{0.98}
\\
\midrule

    \multirow{4}{*}{\textbf{Method}} & \multicolumn{5}{c}{\textbf{Mistral-SFT (7B)}} & \multicolumn{5}{c}{\textbf{Mistral-Instruct (7B)}} \\
    \cmidrule(r){2-6} \cmidrule(lr){7-11}
    &\multicolumn{1}{c}{\textbf{MT-bench}} & \multicolumn{2}{c}{\textbf{AlpacaEval 2}} &\multicolumn{2}{c}{\textbf{HH-RLHF}} & \multicolumn{1}{c}{\textbf{MT-bench}} & \multicolumn{2}{c}{\textbf{AlpacaEval 2}} &\multicolumn{2}{c}{\textbf{HH-RLHF}}\\
    \cmidrule(r){2-2} \cmidrule(lr){3-4} \cmidrule(lr){5-6} \cmidrule(lr){7-7}\cmidrule(lr){8-9} \cmidrule(lr){10-11}
           & \textbf{GPT-4}  & \textbf{LC (\%)} & \textbf{WR~(\%)} & \textbf{Helpful} & \textbf{Harmless} & \textbf{GPT-4}  & \textbf{LC (\%)} & \textbf{WR~(\%)} & \textbf{Helpful} & \textbf{Harmless} \\
\midrule
Base Model & 5.73 & 20.15 & 17.24 & 0.56 & 0.87 & 6.39 & 32.81 & 34.86 & 0.66 & \underline{0.94} \\
\textit{w.} DPO & 5.91 & 31.28 & 32.65 & \underline{0.66} & 0.91 & 6.29 & 35.60 & 37.73 & 0.67 & 0.92 \\
\textit{w.} SimPO & 6.17 & 31.16 & 33.72 & 0.63 & 0.86 & 6.36 & 35.78 & 40.21 & 0.67 & 0.93 \\
\textit{w.} BPO & 5.55 & 18.23 & 17.23 & 0.64 & 0.92 & 5.99 & 19.61 & 27.49 & 0.66 & 0.93 \\
\textit{w.} ARGS & 5.12 & 11.07 & 13.95 & 0.55 & 0.87 & 6.20 & 26.60 & 29.68 & 0.66 & 0.92 \\
\textit{w.} BoN & 6.21 & 33.36 & 27.74 & 0.64 & 0.94 & 6.40 & 34.75 & 39.24 & \underline{0.68} & \underline{0.94} \\
\textit{w.} Aligner & 4.27 & 18.27 & 15.53 & 0.60 & \underline{0.95} & 4.42 & 28.88 & 30.30 & 0.66 & 0.93 \\
\textit{w.} MetaAligner & 4.08 & 12.40 & 9.72 & 0.51 & 0.85 & 3.71 & 18.55 & 16.91 & 0.55 & 0.91 \\
\arrayrulecolor{gray}
\midrule
\arrayrulecolor{black}
\textit{w.} \model-8B & \underline{6.87} & \underline{34.42} & \underline{40.08} & 0.62 & 0.88 & \underline{7.04} & \underline{35.92} & \underline{40.70} & \underline{0.68} & 0.92 \\
\textit{w.} \model-9B &\textbf{7.13} & \textbf{48.99} & \textbf{52.87} & \textbf{0.70} & \textbf{0.96} & \textbf{7.33} & \textbf{56.22} & \textbf{61.71} & \textbf{0.72} & \textbf{0.98}\\
\bottomrule
  \end{tabular}}
  \caption{\textbf{Comparison results with baseline methods.} \model~achieves the superior alignment performance across all benchmarks, outperforming the training-based baselines (i.e., SimPO and DPO) in various settings. Notably, \model~requires only a single training session and is applicable to multiple base models. The best result is highlighted in \textbf{bold}, while the second-best result is highlighted with \underline{underline}.
}
\label{tab:comparison_results}
\end{table*}
\begin{table*}[t]
\centering
\resizebox{.98\textwidth}{!}{
  \begin{tabular}{@{}l
  >{\centering\arraybackslash}m{1.3cm}
  >{\centering\arraybackslash}m{1.3cm}
  >{\centering\arraybackslash}m{1.3cm}
  >{\centering\arraybackslash}m{1.3cm}
  >{\centering\arraybackslash}m{1.3cm}
  >{\centering\arraybackslash}m{1.3cm}
  >{\centering\arraybackslash}m{1.3cm}
  >{\centering\arraybackslash}m{1.7cm}
  @{}}
    \toprule
     \multirow{2}{*}{\textbf{Base Models}} 
    &\multicolumn{3}{c}{\textbf{MT-bench}} & \multicolumn{2}{c}{\textbf{AlpacaEval 2}} &\multicolumn{3}{c}{\textbf{HH-RLHF}} \\
    \cmidrule(r){2-4} \cmidrule(lr){5-6} \cmidrule(lr){7-9} 
           & \textbf{1-Turn} & \textbf{2-Turn} & \textbf{Avg.} & \textbf{LC (\%)} & \textbf{WR~(\%)} & \textbf{Overall} & \textbf{Helpful} & \textbf{Harmless}   \\

\midrule
Llama-3.2-1B-Instruct & 5.32 & 5.13 & 5.25 & 15.57 & 19.09 & 0.0955 & 0.5978 & 0.9313 \\
\textit{w.}~\model-2B & 6.97 & 5.97 & 6.47 & 39.42 & 44.20 & 0.1077 & 0.6948 & 0.9728 \\
\textit{w.}~\model-9B & 7.56 & 6.95 & 7.30 & 50.70 & 56.08 & 0.1130 & 0.7059 & 0.9770 \\
\midrule
Llama-3.2-3B-Instruct & 6.84 & 6.06 & 6.45 & 33.41 & 37.43 & 0.1037 & 0.6533 & 0.9183 \\
\textit{w.}~\model-2B & 7.13 & 6.46 & 6.79 & 39.94 & 45.32 & 0.1069 & 0.6956 & 0.9682 \\
\textit{w.}~\model-9B & 7.58 & 7.00 & 7.36 & 54.06 & 59.30 & 0.1132 & 0.7106 & 0.9875 \\
\midrule
Llama-3-8B-SFT+DPO & 6.70 & 6.48 & 6.59 & 29.84 & 36.77 & 0.1044 & 0.6814 & 0.8900\\
\textit{w.}~\model-9B & 7.42 & 7.03 & 7.22 & 54.29 & 59.51 & 0.1134 & 0.7178 & 0.9765\\
Llama-3-8B-SFT+SimPO & 6.63 & 6.61 & 6.62 & 32.27 & 40.96 & 0.1022 & 0.6640 & 0.8589 \\
\textit{w.}~\model-9B & 7.59 & 7.08 & 7.42 & 55.66 & 60.67 & 0.1121 & 0.7156 & 0.9638\\
\midrule
Llama-3-8B-it+DPO & 6.75 & 7.05 & 6.90 & 47.20 & 53.56 & 0.1120 & 0.7387 & 0.9154 \\
\textit{w.}~\model-9B & 7.79 & 6.98 & 7.39 & 58.56 & 63.60 & 0.1140 & 0.7211 & 0.9831 \\
Llama-3-8B-it+SimPO & 7.09 & 7.00 & 7.05 & 52.57 & 58.33 & 0.1143 & 0.7483 & 0.9182\\
\textit{w.}~\model-9B & 7.43 & 7.22 & 7.33 & 59.66 & 65.32 & 0.1142 & 0.7229 & 0.9756\\
\midrule
Llama-3-70B-Instruct & 7.41 & 7.59 & 7.5 & 46.14 & 51.12 & 0.1087 & 0.6928 & 0.9163  \\
\textit{w.}~\model-9B & 8.23 & 7.28 & 7.75 & 58.18 & 62.34 & 0.1137 & 0.719 & 0.9757  \\
\midrule
Qwen2.5-3B-Instruct & 6.73 & 5.59 & 6.16 & 35.52 & 40.42 & 0.1050 & 0.6802 & 0.9587 \\
\textit{w.}~\model-2B & 7.06 & 6.26 & 6.66 & 42.63 & 47.83 & 0.1065 & 0.6973 & 0.9771 \\
\textit{w.}~\model-9B & 7.58 & 7.24 & 7.41 & 55.71 & 61.43 & 0.1132 & 0.7106 & 0.9875 \\
\midrule
Qwen2.5-7B-Instruct & 7.11 & 6.96 & 7.03 & 45.03 & 49.95 & 0.1095 & 0.6995 & 0.9442 \\
\textit{w.}~\model-2B & 7.01 & 6.34 & 6.67 & 43.89 & 49.07 & 0.1074 & 0.7013 & 0.9659 \\
\textit{w.}~\model-9B & 7.62 & 7.10 & 7.35 & 57.89 & 63.01 & 0.1117 & 0.7100 & 0.9445 \\
\midrule
Qwen2.5-14B-Instruct & 7.24 & 6.71 & 6.98 & 51.60 & 57.10 & 0.1117 & 0.7100 & 0.9445 \\
\textit{w.}~\model-2B & 7.08 & 6.33 & 6.71 & 43.70 & 48.76 & 0.1078 & 0.7017 & 0.9704 \\
\textit{w.}~\model-9B & 7.62 & 7.35 & 7.48 & 55.13 & 60.65 & 0.1136 & 0.7185 & 0.9759 \\
\midrule
Qwen2.5-32B-Instruct & 7.35 & 6.95 & 7.15 & 54.93 & 60.95 & 0.1132 & 0.7189 & 0.9594 \\
\textit{w.}~\model-9B & 7.58 & 7.63 & 7.60 & 55.13 & 61.94 & 0.1143 & 0.7248 & 0.9797 
\\
\midrule
GPT-4o (API) & 7.40 & 7.47 & 7.43 & 53.64 & 62.01 & 0.1119 & 0.6974 & 0.9669 \\
\textit{w.}~\model-9B & 7.66 & 7.37 & 7.51 & 58.91 & 64.30 & 0.1129 & 0.7167 & 0.9893 
\\

\arrayrulecolor{black}
\bottomrule
  \end{tabular}}
  \caption{\textbf{Performance of \model~models.} The results demonstrate that both \model-2B and \model-9B lead to performance improvements across different base models. \model-9B enhances the performance of larger models, such as Qwen2.5-14B and 32B, as well as black-box GPT-4o, exhibiting a weak-to-strong improvement pattern. Furthermore, the results show that \model~can be effectively integrated with existing preference optimization methods, such as DPO and SimPO, further enhancing alignment performance.}
  \label{tab:various_base_models}
\end{table*}

\subsection{Experiment Setup}

\paragraph{Evaluation Benchmarks and Metrics.} 
We conduct our experiments using two widely recognized benchmarks for open-ended instruction-following: \textit{MT-Bench}~\cite{zheng2023judging} and \textit{AlpacaEval 2}~\cite{dubois2024length}. These benchmarks are designed to evaluate the conversational abilities of models across a diverse set of queries. \textit{AlpacaEval 2} includes 805 questions drawn from five distinct datasets, while \textit{MT-Bench} covers eight categories and comprises a total of 80 questions.
Additionally, we employ the \textit{HH-RLHF}~\cite{bai2022training} datasets to assess how well the models’ generative capabilities align with human values, particularly emphasizing helpfulness and harmlessness. 
We adhere to each benchmark's specific evaluation protocol to report scores. In \textit{AlpacaEval 2}, we report both the raw win rate (WR) and the length-controlled win rate (LC), comparing performance against the GPT-4 model. In contrast, we present the average score for \textit{MT-Bench}, also utilizing GPT-4 as the judge model.
For \textit{HH-RLHF}, we report scores that reflect the models' helpfulness and harmlessness, as well as the overall score. These scores are measured using \textit{ArmoRM}~\cite{ArmoRM}, a state-of-the-art reward model from \textit{RewardBench}~\cite{lambert2024rewardbench}, designed to align with human preferences.

\begin{figure*}[t] 
    \centering
    \includegraphics[width=1\textwidth]{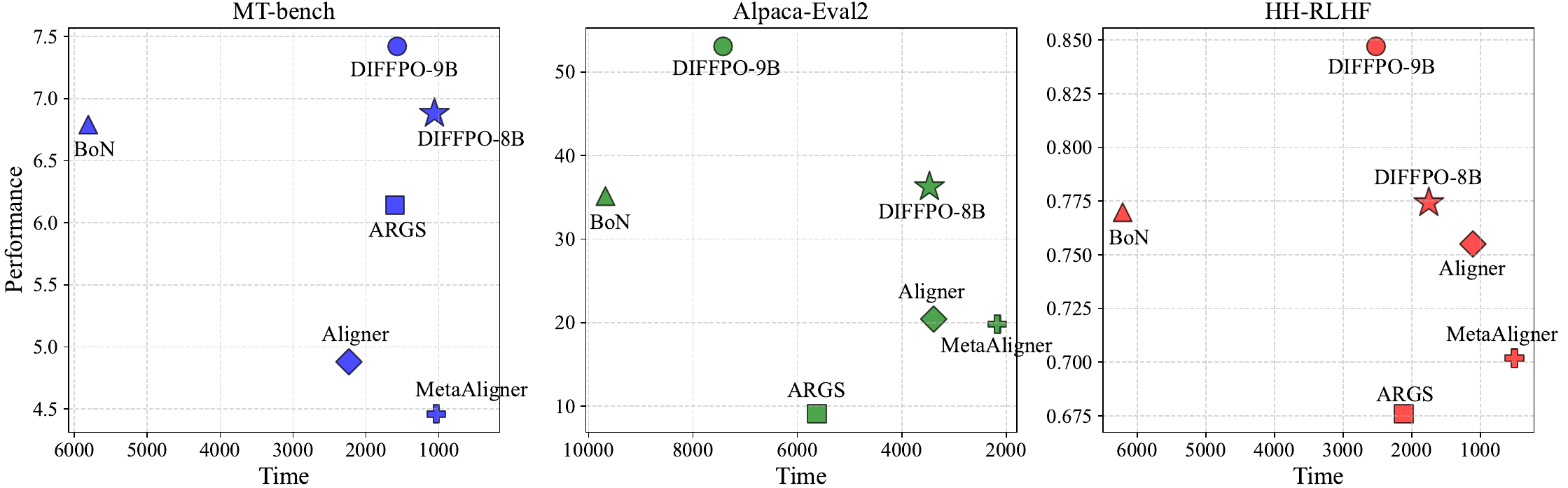}
    \caption{\textbf{Comparison of Inference-Time Efficiency.} We compare \model~with existing inference-time alignment techniques, evaluating both alignment performance and execution time. Points located closer to the \textbf{\textit{top-right}} corner indicate a better trade-off. When considering both aspects, \model~demonstrates a surpassing performance-efficiency trade-off on all three datasets.}
    \label{fig: performance_vs_time}
    \vspace{-6pt}
\end{figure*}

\vspace{-3pt}
\paragraph{Baselines.}
We compare \model~with two primary categories of offline preference optimization methods. In the category of \textbf{training-based} methods:
\textit{Direct Preference Optimization (DPO)}~\cite{rafailov2024direct} reparameterizes reward functions to simplify and stabilize the preference learning process.
\textit{SimPO}~\cite{meng2024simposimplepreferenceoptimization} utilizes the average log probability of a sequence as an implicit reward, aligning more closely with model generation.
For \textbf{training-free} methods: \textit{Black-Box Prompt Optimization (BPO)}~\cite{cheng2024blackboxpromptoptimizationaligning} adapts user prompts to better align with LLMs' input comprehension, achieving user intents optimally without altering LLM parameters. \textit{ARGS}~\cite{khanov2024argsalignmentrewardguidedsearch} integrates alignment into the decoding process through reward-guided search, eliminating the need for costly RL training. \textit{Best-of-N sampling (BoN)}~\cite{nakano2021webgpt} samples N times and selects the highest-scoring sample based on the reward model, with N set to 4 in our experiments using \textit{ArmoRM}~\cite{ArmoRM} as the reward model. 
Furthermore, \textit{Aligner}~\cite{ji2024aligner} and \textit{MetaAligner}~\cite{yang2024metaaligner} employ an additional model to learn corrective residuals between preferred and dispreferred responses to refine model generation.

\begin{table}[t]
\centering
\resizebox{0.48\textwidth}{!}{
\begin{tabular}{@{}lrrrrr@{}}  
\toprule
\textbf{Block Size} & {\textbf{16}} & {\textbf{32}} & {\textbf{64}} & {\textbf{128}} & {\textbf{256}} \\
\midrule
\textbf{MT (GPT-4)} & 6.86 & \textbf{6.96} & 6.84 & 6.81 & 6.77 \\
\textbf{Time (s)} & 1080 & \textbf{1012} & 1390 & 1520 & 1937 \\
\midrule
\textbf{AE2 (LC)} & 33.52 & 33.54 & 33.46 & 32.98 & \textbf{36.24} \\
\textbf{Time (s)} & 3712 & \textbf{3471} & 4614 & 5510 & 7520 \\
\midrule
\textbf{HH (Avg.)} & 0.7742 & 0.7749 & 0.7743 & 0.7741 & \textbf{0.7761} \\
\textbf{Time (s)} & 1564 & \textbf{1551} & 1620 & 1816 & 2684 \\
\bottomrule
\end{tabular}}
\caption{\textbf{Performance Under Hybrid Decoding.}
We segment the vanilla generation into blocks of varying sizes and sequentially apply \model-8B to each block. This approach allows \model~decoding to be parallel within blocks and auto-regressive between blocks. Hybrid decoding significantly reduces the decoding time, indicating a feasible trade-off for performance.}
\label{tab:Hybrid Decoding Strategy}
\vspace{-6pt}
\end{table}

\vspace{-3pt}
\paragraph{Base Models and Inference Settings.} 
We perform preference optimization primarily on two model families: \textit{Llama-3-8B}~\cite{llama3modelcard} and \textit{Mistral-7B}~\cite{jiang2023mistral}, under two configurations: SFT and Instruct. 
In the SFT configuration, we utilize open-source models from \textit{SimPO}~\cite{meng2024simposimplepreferenceoptimization} that follow \textit{Zephyr}~\cite{tunstall2023zephyr} to train the base models (i.e., \texttt{meta-llama/Meta-Llama-3-8B}) on the \textit{UltraChat-200k}~\cite{ding2023enhancing} dataset to derive an SFT model.
For the \text{Instruct configuration}, we employ an off-the-shelf instruction-tuned models (i.e., \texttt{meta-llama/Meta-Llama-3-8B-Instruct}).
To further validate scalability, we conduct additional experiments using the \textit{Llama-3.2} series, \textit{Qwen-2.5} series~\cite{qwen2.5}, and GPT-4o~\cite{achiam2023gpt} as the base models.

During the inference phase of \model, we initially generate responses using the base models. For each benchmark. In \textit{AlpacaEval 2} and \textit{HH-RLHF}, we employ a sampling decoding strategy with a temperature setting of 0.7.
For \textit{MT-Bench}, we adhere to the official decoding configuration, which specifies varying temperatures for different categories.
In our primary experiments, we set the maximum token generation length to 256. Results for experiments conducted at various lengths are provided in Tab.~\ref{Scaling Across Various Context Lengths}. Subsequently, the responses generated by the base models are aligned using the trained \model. For the main results, parallel decoding is executed with a block size of 256.


\subsection{Experiment Results}

\paragraph{\model~significantly outperforms existing preference optimization methods.}
As shown in Table~\ref{tab:comparison_results}, while all preference optimization algorithms improve performance over the base model, \model~achieves the best overall performance across all benchmarks and settings. These consistent and significant improvements underscore the robustness and effectiveness of \model. Notably, \model~outperforms the training-based baselines (i.e., SimPO and DPO) across various settings, despite requiring only a single training session of \model~model and being capable of enhancing the performance of multiple base models.


\paragraph{\model~consistently improves the performance of base models of various sizes.}
We report the performance of \model-2B and \model-9B on base models of various sizes, with the results presented in Table~\ref{tab:various_base_models}. The results demonstrate that both \model-2B and \model-9B lead to performance improvements across different base models. However, the performance gain of \model-2B is limited, showing notable improvements primarily for smaller models. In contrast, \model-9B enhances the performance of larger models, such as Qwen2.5-14B and 32B, as well as black-box GPT-4, exhibiting a weak-to-strong improvement pattern. Furthermore, the results show that \model~can be effectively integrated with existing preference optimization methods, such as DPO and SimPO, further enhancing alignment performance. These results underscore the scalability of \model.

\paragraph{\model~achieves a surpassing performance-efficiency trade-off.}
We compare \model~with existing inference-time alignment techniques, evaluating both alignment performance and execution time. The results are illustrated in Fig.~\ref{fig: performance_vs_time}, with the execution time measured on a single NVIDIA A100 80GB GPU. Points located closer to the top-right corner indicate a more favorable Pareto frontier. BoN and MetaAligner achieves commendable alignment performance and inference time respectively. However, when considering both aspects, \model~demonstrates a surpassing performance-efficiency trade-off on all three datasets. The experiments are conducted on Llama-3-SFT.

\subsection{Analysis}

\paragraph{Performance Under Hybrid Decoding.}
We investigate the hybrid decoding strategy of \model, with results provided in Tab~\ref{tab:Hybrid Decoding Strategy}. We segment the vanilla generation, which has a maximum length of 256, into blocks of varying sizes and sequentially apply \model-8B to each block. This approach allows \model~decoding to be parallel within blocks and auto-regressive between blocks. It can be observed that hybrid decoding significantly reduces the decoding time, with optimal efficiency achieved at a block size of 32. On the other hand, performance is enhanced when the block size is set to 256, which corresponds to purely parallel decoding, indicating a feasible trade-off. The experiments are conducted on Llama-3-SFT.

\vspace{-3pt}
\paragraph{Scaling towards Longer Generation Lengths.}
We validate the scalability of the \model~model in response to increasing generation lengths, with results presented in Tab~\ref{Scaling Across Various Context Lengths}. Using base models, we generate outputs on \textit{MT-Bench} under various maximum length settings and observe a positive correlation between increased text length and higher scores. Subsequently, the same optimized \model-8B and 9B is applied to these outputs using the hybrid decoding strategy described in the previous section. This approach consistently yields enhanced alignment performance, demonstrating \model's robust scaling capabilities towards longer generation lengths.

\vspace{-3pt}
\paragraph{Loss and Hyperparameter Ablation.} 
We evaluate the effectiveness of the training loss of \model~in Section~\ref{sec:Consistency Optimization} and the inference strategy in Section~\ref{sec:Practical Implementations}. The results are presented in Table~\ref{tab:Loss and Hyperparameter Ablation}. We report on two decoding strategies: \textit{vanilla decoding} of a single model and \textit{\model~decoding}, which applies the optimized \model-9B on the output of the base model. The findings indicate that applying \model~to the base model achieves performance superior to that of single models alone, thus demonstrating the effectiveness of the \model~strategy.
Furthermore, we report the results of an ablation study on the hyperparameter $w$ in Eq.~\ref{total_loss}. When using \model~decoding, employing $L_{\mathrm{Con}}$ with larger values of $w$ lead to a more pronounced improvement in performance.

\section{Conclusion}
This paper introduces a novel inference-time alignment framework for large language models, \model. \model~achieves alignment at the sentence level to better model human preferences, drawing inspiration from the denoising process. \model~outperforms both strong training-based and inference-time alignment techniques in terms of alignment performance and inference speed. Experiments scaling \model~from 2B to 9B parameters, expanding the base model from 1B to 70B, and increasing the context length from 256 to 2,048 demonstrate that \model~is a robust and scalable framework for LLM alignment.

\begin{table}[H]
\centering
\resizebox{0.48\textwidth}{!}{
\begin{tabular}{@{}l>{\centering\arraybackslash}m{1.0cm}
>{\centering\arraybackslash}m{1.0cm}
>{\centering\arraybackslash}m{1.0cm}
>{\centering\arraybackslash}m{1.0cm}@{}}
\toprule
\textbf{Generation Length} & 256 & 512 & 1,024 & 2,048 \\
\midrule
Llama-3-SFT & 6.21 & 6.61 & 6.76 & 6.71 \\
\textit{w.} \model-8B ($\Delta$) & \textcolor{gray}{+0.75} & \textcolor{gray}{+0.81} & \textcolor{gray}{+0.93} & \textcolor{gray}{+1.05} \\
\textit{w.} \model-9B ($\Delta$) & \textcolor{gray}{+1.24} & \textcolor{gray}{+1.64} & \textcolor{gray}{+1.48} & \textcolor{gray}{+0.50} \\
\arrayrulecolor{gray}
\midrule
Llama-3-Instruct & 6.78 & 7.87 & 7.99 & 8.00 \\
\textit{w.} \model-8B ($\Delta$) & \textcolor{gray}{+0.24} & \textcolor{gray}{-0.12} & \textcolor{gray}{+0.01} & \textcolor{gray}{+0.02} \\
\textit{w.} \model-9B ($\Delta$) & \textcolor{gray}{+0.62} & \textcolor{gray}{+0.68} & \textcolor{gray}{+0.34} & \textcolor{gray}{+0.62} \\
\midrule
Mistral-SFT & 5.73 & 6.42 & 6.50 & 6.36 \\
\textit{w.} \model-8B ($\Delta$) & \textcolor{gray}{+1.14} & \textcolor{gray}{+1.16} & \textcolor{gray}{+1.25} & \textcolor{gray}{+1.51} \\
\textit{w.} \model-9B ($\Delta$) & \textcolor{gray}{+1.40} & \textcolor{gray}{+1.47} & \textcolor{gray}{+1.71} & \textcolor{gray}{+1.86} \\
\midrule
Mistral-Instruct & 6.39 & 7.47 & 7.68 & 7.64 \\
\textit{w.} \model-8B ($\Delta$) & \textcolor{gray}{+0.65} & \textcolor{gray}{+0.06} & \textcolor{gray}{+0.13} & \textcolor{gray}{+0.17} \\
\textit{w.} \model-9B ($\Delta$) & \textcolor{gray}{+0.94} & \textcolor{gray}{+0.59} & \textcolor{gray}{+0.58} & \textcolor{gray}{+0.78} \\
\arrayrulecolor{black}
\bottomrule
\end{tabular}
}
\caption{\textbf{Scaling towards Longer Generation Lengths.} We evaluate the performance of \model \  under various maximum length settings. When the same optimized \model-8B and 9B is applied to these outputs, consistently enhanced performance demonstrates \model's robust scaling capabilities.}
\label{Scaling Across Various Context Lengths}
\vspace{1em}
\centering
\resizebox{.48\textwidth}{!}{
  \begin{tabular}{@{}lccc
  @{}}
    \toprule
     \multirow{2}{*}{\textbf{Loss}} 
    &\multicolumn{1}{c}{\textbf{MT}} & \multicolumn{1}{c}{\textbf{AE2}} &\multicolumn{1}{c}{\textbf{HH}} \\
    \cmidrule(r){2-2} \cmidrule(lr){3-3} \cmidrule(lr){4-4} 
           & \multicolumn{1}{c}{\textbf{GPT-4}} & \multicolumn{1}{c}{\textbf{LC (\%)}} &\multicolumn{1}{c}{\textbf{Avg.}}  \\
           \midrule
           \multicolumn{4}{c}{\textit{Vanilla Decoding}} \\
           Llama-3-Instruct&6.78 & 36.83 & 0.7985\\ 
           $L_{\mathrm{AR}}$& 6.90 & 36.35 & 0.7971\\
           \midrule
           \multicolumn{4}{c}{\textit{Llama-3-it} \textit{w.} \model-8B} \\
           $L_{\mathrm{AR}}$ only& 6.75 & 35.84 & 0.7968\\
$10 \times L_{\mathrm{Con}} + L_{\mathrm{AR}}$& 6.85 & 35.96 & 0.7997\\
$100 \times L_{\mathrm{Con}} + L_{\mathrm{AR}}$& 6.86 & 35.92 & 0.7998\\
$1,000 \times L_{\mathrm{Con}} + L_{\mathrm{AR}}$& 7.02 & 36.44 & 0.7998\\
\bottomrule
  \end{tabular}}
  \caption{\textbf{Loss and Hyperparameter Ablation.} We report the results of vanilla decoding from the base model and the optimized \model~model. The results indicate that applying \model~to the base model yields outperforming performance than single models, demonstrating the effectiveness of \model~strategy.}
\label{tab:Loss and Hyperparameter Ablation}
\end{table}

\clearpage

\section*{Limitations}
We acknowledge the presence of certain limitations.
While \model\ has demonstrated a superior trade-off between performance and inference-time cost, it still introduces additional inference latency due to the need for an extra model for alignment. Moreover, we observe that the performance of \model\ scales with its size, which presents challenges for cost-effectiveness during deployment. Additionally, despite the empirical success and intuitive motivation behind \model, a more rigorous theoretical analysis is required to fully understand its effectiveness. Future work could explore how to combine the diffusion process (i.e., the denoising process) with the alignment task more effectively. This paper draws insights from the analogy between the denoising process and alignment. We hope our findings will facilitate future exploration of existing successful techniques in the natural language processing domain.

\section*{Potential Risks}
As an inference-time alignment technique, \model \ aims to develop AI assistants that align with positive human intentions and social values. However, there is a potential risk that \model\ could be misused to align with harmful or negative values. We strongly oppose any such misuse, as it could hinder human progress, and advocate for the responsible and ethical use of \model.

\section*{Acknowledgements}
This work is supported by the National Natural Science Foundation of China (Grant No. 12326612, 62476241), the Natural Science Foundation of Zhejiang Province, China (Grant No. LZ23F020008), and the Zhejiang University-Angelalign Inc. R\&D Center for Intelligent Healthcare.

\newpage

\bibliography{acl_latex} 

\begin{thebibliography}{66}
\providecommand{\natexlab}[1]{#1}

\bibitem[{Achiam et~al.(2023)Achiam, Adler, Agarwal, Ahmad, Akkaya, Aleman, Almeida, Altenschmidt, Altman, Anadkat et~al.}]{achiam2023gpt}
Josh Achiam, Steven Adler, Sandhini Agarwal, Lama Ahmad, Ilge Akkaya, Florencia~Leoni Aleman, Diogo Almeida, Janko Altenschmidt, Sam Altman, Shyamal Anadkat, et~al. 2023.
\newblock Gpt-4 technical report.
\newblock \emph{arXiv preprint arXiv:2303.08774}.

\bibitem[{Ahmadian et~al.(2024)Ahmadian, Cremer, Gall{\'e}, Fadaee, Kreutzer, Pietquin, {\"U}st{\"u}n, and Hooker}]{ahmadian2024back}
Arash Ahmadian, Chris Cremer, Matthias Gall{\'e}, Marzieh Fadaee, Julia Kreutzer, Olivier Pietquin, Ahmet {\"U}st{\"u}n, and Sara Hooker. 2024.
\newblock Back to basics: Revisiting reinforce style optimization for learning from human feedback in llms.
\newblock \emph{arXiv preprint arXiv:2402.14740}.

\bibitem[{AI@Meta(2024)}]{llama3modelcard}
AI@Meta. 2024.
\newblock \href {https://github.com/meta-llama/llama3/blob/main/MODEL_CARD.md} {Llama 3 model card}.

\bibitem[{Andrychowicz et~al.(2017)Andrychowicz, Wolski, Ray, Schneider, Fong, Welinder, McGrew, Tobin, Pieter~Abbeel, and Zaremba}]{andrychowicz2017hindsight}
Marcin Andrychowicz, Filip Wolski, Alex Ray, Jonas Schneider, Rachel Fong, Peter Welinder, Bob McGrew, Josh Tobin, OpenAI Pieter~Abbeel, and Wojciech Zaremba. 2017.
\newblock Hindsight experience replay.
\newblock \emph{Advances in neural information processing systems}, 30.

\bibitem[{Bai et~al.(2022)Bai, Jones, Ndousse, Askell, Chen, DasSarma, Drain, Fort, Ganguli, Henighan et~al.}]{bai2022training}
Yuntao Bai, Andy Jones, Kamal Ndousse, Amanda Askell, Anna Chen, Nova DasSarma, Dawn Drain, Stanislav Fort, Deep Ganguli, Tom Henighan, et~al. 2022.
\newblock Training a helpful and harmless assistant with reinforcement learning from human feedback.
\newblock \emph{arXiv preprint arXiv:2204.05862}.

\bibitem[{Chakraborty et~al.(2024)Chakraborty, Ghosal, Yin, Manocha, Wang, Bedi, and Huang}]{chakraborty2024transfer}
Souradip Chakraborty, Soumya~Suvra Ghosal, Ming Yin, Dinesh Manocha, Mengdi Wang, Amrit~Singh Bedi, and Furong Huang. 2024.
\newblock Transfer q star: Principled decoding for llm alignment.
\newblock \emph{arXiv preprint arXiv:2405.20495}.

\bibitem[{Chen et~al.(2023)Chen, Borgeaud, Irving, Lespiau, Sifre, and Jumper}]{chen2023accelerating}
Charlie Chen, Sebastian Borgeaud, Geoffrey Irving, Jean-Baptiste Lespiau, Laurent Sifre, and John Jumper. 2023.
\newblock Accelerating large language model decoding with speculative sampling.
\newblock \emph{arXiv preprint arXiv:2302.01318}.

\bibitem[{Chen et~al.(2024{\natexlab{a}})Chen, Hu, Feng, and Liu}]{chen2024learnable}
Ruizhe Chen, Tianxiang Hu, Yang Feng, and Zuozhu Liu. 2024{\natexlab{a}}.
\newblock Learnable privacy neurons localization in language models.
\newblock \emph{arXiv preprint arXiv:2405.10989}.

\bibitem[{Chen et~al.(2024{\natexlab{b}})Chen, Li, Yang, Zhou, and Liu}]{chen2024editable}
Ruizhe Chen, Yichen Li, Jianfei Yang, Joey~Tianyi Zhou, and Zuozhu Liu. 2024{\natexlab{b}}.
\newblock Editable fairness: Fine-grained bias mitigation in language models.
\newblock \emph{arXiv preprint arXiv:2408.11843}.

\bibitem[{Chen et~al.(2024{\natexlab{c}})Chen, Yang, Xiong, Bai, Hu, Hao, Feng, Zhou, Wu, and Liu}]{chen2024fast}
Ruizhe Chen, Jianfei Yang, Huimin Xiong, Jianhong Bai, Tianxiang Hu, Jin Hao, Yang Feng, Joey~Tianyi Zhou, Jian Wu, and Zuozhu Liu. 2024{\natexlab{c}}.
\newblock Fast model debias with machine unlearning.
\newblock \emph{Advances in Neural Information Processing Systems}, 36.

\bibitem[{Chen et~al.(2024{\natexlab{d}})Chen, Zhang, Luo, Chai, and Liu}]{chen2024pad}
Ruizhe Chen, Xiaotian Zhang, Meng Luo, Wenhao Chai, and Zuozhu Liu. 2024{\natexlab{d}}.
\newblock Pad: Personalized alignment of llms at decoding-time.
\newblock \emph{arXiv preprint arXiv:2410.04070}.

\bibitem[{Cheng et~al.(2024)Cheng, Liu, Zheng, Ke, Wang, Dong, Tang, and Huang}]{cheng2024blackboxpromptoptimizationaligning}
Jiale Cheng, Xiao Liu, Kehan Zheng, Pei Ke, Hongning Wang, Yuxiao Dong, Jie Tang, and Minlie Huang. 2024.
\newblock \href {https://arxiv.org/abs/2311.04155} {Black-box prompt optimization: Aligning large language models without model training}.
\newblock \emph{Preprint}, arXiv:2311.04155.

\bibitem[{Christiano et~al.(2017)Christiano, Leike, Brown, Martic, Legg, and Amodei}]{christiano2017deep}
Paul~F Christiano, Jan Leike, Tom Brown, Miljan Martic, Shane Legg, and Dario Amodei. 2017.
\newblock Deep reinforcement learning from human preferences.
\newblock \emph{Advances in neural information processing systems}, 30.

\bibitem[{Cui et~al.(2023)Cui, Yuan, Ding, Yao, Zhu, Ni, Xie, Liu, and Sun}]{cui2023ultrafeedback}
Ganqu Cui, Lifan Yuan, Ning Ding, Guanming Yao, Wei Zhu, Yuan Ni, Guotong Xie, Zhiyuan Liu, and Maosong Sun. 2023.
\newblock \href {https://arxiv.org/abs/2310.01377} {Ultrafeedback: Boosting language models with high-quality feedback}.
\newblock \emph{Preprint}, arXiv:2310.01377.

\bibitem[{Dhariwal and Nichol(2021)}]{dhariwal2021diffusion}
Prafulla Dhariwal and Alexander Nichol. 2021.
\newblock Diffusion models beat gans on image synthesis.
\newblock \emph{Advances in neural information processing systems}, 34:8780--8794.

\bibitem[{Ding et~al.(2023)Ding, Chen, Xu, Qin, Zheng, Hu, Liu, Sun, and Zhou}]{ding2023enhancing}
Ning Ding, Yulin Chen, Bokai Xu, Yujia Qin, Zhi Zheng, Shengding Hu, Zhiyuan Liu, Maosong Sun, and Bowen Zhou. 2023.
\newblock Enhancing chat language models by scaling high-quality instructional conversations.
\newblock \emph{arXiv preprint arXiv:2305.14233}.

\bibitem[{Dubois et~al.(2024)Dubois, Galambosi, Liang, and Hashimoto}]{dubois2024length}
Yann Dubois, Bal{\'a}zs Galambosi, Percy Liang, and Tatsunori~B Hashimoto. 2024.
\newblock Length-controlled alpacaeval: A simple way to debias automatic evaluators.
\newblock \emph{arXiv preprint arXiv:2404.04475}.

\bibitem[{Ethayarajh et~al.(2024)Ethayarajh, Xu, Muennighoff, Jurafsky, and Kiela}]{ethayarajh2024kto}
Kawin Ethayarajh, Winnie Xu, Niklas Muennighoff, Dan Jurafsky, and Douwe Kiela. 2024.
\newblock Kto: Model alignment as prospect theoretic optimization.
\newblock \emph{arXiv preprint arXiv:2402.01306}.

\bibitem[{Fan et~al.(2024{\natexlab{a}})Fan, Chen, Hu, and Liu}]{fan2024fairmt}
Zhiting Fan, Ruizhe Chen, Tianxiang Hu, and Zuozhu Liu. 2024{\natexlab{a}}.
\newblock Fairmt-bench: Benchmarking fairness for multi-turn dialogue in conversational llms.
\newblock \emph{arXiv preprint arXiv:2410.19317}.

\bibitem[{Fan et~al.(2024{\natexlab{b}})Fan, Chen, Xu, and Liu}]{fan2024biasalert}
Zhiting Fan, Ruizhe Chen, Ruiling Xu, and Zuozhu Liu. 2024{\natexlab{b}}.
\newblock Biasalert: A plug-and-play tool for social bias detection in llms.
\newblock \emph{arXiv preprint arXiv:2407.10241}.

\bibitem[{Fu et~al.(2024)Fu, Bailis, Stoica, and Zhang}]{fu2024break}
Yichao Fu, Peter Bailis, Ion Stoica, and Hao Zhang. 2024.
\newblock Break the sequential dependency of llm inference using lookahead decoding.
\newblock \emph{arXiv preprint arXiv:2402.02057}.

\bibitem[{Gong et~al.(2022)Gong, Li, Feng, Wu, and Kong}]{gong2022diffuseq}
Shansan Gong, Mukai Li, Jiangtao Feng, Zhiyong Wu, and LingPeng Kong. 2022.
\newblock Diffuseq: Sequence to sequence text generation with diffusion models.
\newblock \emph{arXiv preprint arXiv:2210.08933}.

\bibitem[{Gong et~al.(2023)Gong, Li, Feng, Wu, and Kong}]{gong2023diffuseq}
Shansan Gong, Mukai Li, Jiangtao Feng, Zhiyong Wu, and Lingpeng Kong. 2023.
\newblock Diffuseq-v2: Bridging discrete and continuous text spaces for accelerated seq2seq diffusion models.
\newblock \emph{arXiv preprint arXiv:2310.05793}.

\bibitem[{Gulrajani and Hashimoto(2024)}]{gulrajani2024likelihood}
Ishaan Gulrajani and Tatsunori~B Hashimoto. 2024.
\newblock Likelihood-based diffusion language models.
\newblock \emph{Advances in Neural Information Processing Systems}, 36.

\bibitem[{Han et~al.(2024)Han, Shenfeld, Srivastava, Kim, and Agrawal}]{han2024value}
Seungwook Han, Idan Shenfeld, Akash Srivastava, Yoon Kim, and Pulkit Agrawal. 2024.
\newblock Value augmented sampling for language model alignment and personalization.
\newblock \emph{arXiv preprint arXiv:2405.06639}.

\bibitem[{Han et~al.(2022)Han, Kumar, and Tsvetkov}]{han2022ssd}
Xiaochuang Han, Sachin Kumar, and Yulia Tsvetkov. 2022.
\newblock Ssd-lm: Semi-autoregressive simplex-based diffusion language model for text generation and modular control.
\newblock \emph{arXiv preprint arXiv:2210.17432}.

\bibitem[{Ho et~al.(2020)Ho, Jain, and Abbeel}]{ho2020denoising}
Jonathan Ho, Ajay Jain, and Pieter Abbeel. 2020.
\newblock Denoising diffusion probabilistic models.
\newblock \emph{Advances in neural information processing systems}, 33:6840--6851.

\bibitem[{Huang et~al.(2024)Huang, Sengupta, Bonadiman, Lai, Gupta, Pappas, Mansour, Kirchhoff, and Roth}]{huang2024deal}
James~Y Huang, Sailik Sengupta, Daniele Bonadiman, Yi-an Lai, Arshit Gupta, Nikolaos Pappas, Saab Mansour, Katrin Kirchhoff, and Dan Roth. 2024.
\newblock Deal: Decoding-time alignment for large language models.
\newblock \emph{arXiv preprint arXiv:2402.06147}.

\bibitem[{Ji et~al.(2024)Ji, Chen, Lou, Hong, Zhang, Pan, Dai, Qiu, and Yang}]{ji2024aligner}
Jiaming Ji, Boyuan Chen, Hantao Lou, Donghai Hong, Borong Zhang, Xuehai Pan, Juntao Dai, Tianyi Qiu, and Yaodong Yang. 2024.
\newblock Aligner: Efficient alignment by learning to correct.
\newblock \emph{arXiv preprint arXiv:2402.02416}.

\bibitem[{Jiang et~al.(2023)Jiang, Sablayrolles, Mensch, Bamford, Chaplot, Casas, Bressand, Lengyel, Lample, Saulnier et~al.}]{jiang2023mistral}
Albert~Q Jiang, Alexandre Sablayrolles, Arthur Mensch, Chris Bamford, Devendra~Singh Chaplot, Diego de~las Casas, Florian Bressand, Gianna Lengyel, Guillaume Lample, Lucile Saulnier, et~al. 2023.
\newblock Mistral 7b.
\newblock \emph{arXiv preprint arXiv:2310.06825}.

\bibitem[{Khanov et~al.(2024)Khanov, Burapacheep, and Li}]{khanov2024argsalignmentrewardguidedsearch}
Maxim Khanov, Jirayu Burapacheep, and Yixuan Li. 2024.
\newblock \href {https://arxiv.org/abs/2402.01694} {Args: Alignment as reward-guided search}.
\newblock \emph{Preprint}, arXiv:2402.01694.

\bibitem[{Kou et~al.(2024)Kou, Hu, He, Deng, and Zhang}]{kou2024cllms}
Siqi Kou, Lanxiang Hu, Zhezhi He, Zhijie Deng, and Hao Zhang. 2024.
\newblock Cllms: Consistency large language models.
\newblock \emph{arXiv preprint arXiv:2403.00835}.

\bibitem[{Lambert et~al.(2024)Lambert, Pyatkin, Morrison, Miranda, Lin, Chandu, Dziri, Kumar, Zick, Choi et~al.}]{lambert2024rewardbench}
Nathan Lambert, Valentina Pyatkin, Jacob Morrison, LJ~Miranda, Bill~Yuchen Lin, Khyathi Chandu, Nouha Dziri, Sachin Kumar, Tom Zick, Yejin Choi, et~al. 2024.
\newblock Rewardbench: Evaluating reward models for language modeling.
\newblock \emph{arXiv preprint arXiv:2403.13787}.

\bibitem[{Leviathan et~al.(2023)Leviathan, Kalman, and Matias}]{leviathan2023fast}
Yaniv Leviathan, Matan Kalman, and Yossi Matias. 2023.
\newblock Fast inference from transformers via speculative decoding.
\newblock In \emph{International Conference on Machine Learning}, pages 19274--19286. PMLR.

\bibitem[{Li et~al.(2022)Li, Thickstun, Gulrajani, Liang, and Hashimoto}]{li2022diffusion}
Xiang Li, John Thickstun, Ishaan Gulrajani, Percy~S Liang, and Tatsunori~B Hashimoto. 2022.
\newblock Diffusion-lm improves controllable text generation.
\newblock \emph{Advances in Neural Information Processing Systems}, 35:4328--4343.

\bibitem[{Li et~al.(2023{\natexlab{a}})Li, Wei, Zhao, Zhang, and Zhang}]{li2023rain}
Yuhui Li, Fangyun Wei, Jinjing Zhao, Chao Zhang, and Hongyang Zhang. 2023{\natexlab{a}}.
\newblock Rain: Your language models can align themselves without finetuning.
\newblock \emph{arXiv preprint arXiv:2309.07124}.

\bibitem[{Li et~al.(2023{\natexlab{b}})Li, Xu, Zhang, Lin, Yu, Sun, and Luo}]{li2023remax}
Ziniu Li, Tian Xu, Yushun Zhang, Zhihang Lin, Yang Yu, Ruoyu Sun, and Zhi-Quan Luo. 2023{\natexlab{b}}.
\newblock Remax: A simple, effective, and efficient reinforcement learning method for aligning large language models.
\newblock In \emph{Forty-first International Conference on Machine Learning}.

\bibitem[{Lightman et~al.(2023)Lightman, Kosaraju, Burda, Edwards, Baker, Lee, Leike, Schulman, Sutskever, and Cobbe}]{lightman2023let}
Hunter Lightman, Vineet Kosaraju, Yura Burda, Harri Edwards, Bowen Baker, Teddy Lee, Jan Leike, John Schulman, Ilya Sutskever, and Karl Cobbe. 2023.
\newblock Let's verify step by step.
\newblock \emph{arXiv preprint arXiv:2305.20050}.

\bibitem[{Lovelace et~al.(2024)Lovelace, Kishore, Wan, Shekhtman, and Weinberger}]{lovelace2024latent}
Justin Lovelace, Varsha Kishore, Chao Wan, Eliot Shekhtman, and Kilian~Q Weinberger. 2024.
\newblock Latent diffusion for language generation.
\newblock \emph{Advances in Neural Information Processing Systems}, 36.

\bibitem[{Luo et~al.(2024)Luo, Deng, Chen, and Liu}]{luo2024faintbench}
Hanjun Luo, Ziye Deng, Ruizhe Chen, and Zuozhu Liu. 2024.
\newblock Faintbench: A holistic and precise benchmark for bias evaluation in text-to-image models.
\newblock \emph{arXiv preprint arXiv:2405.17814}.

\bibitem[{Lyu et~al.(2023)Lyu, Luo, Shi, Hollon, and Lee}]{lyu2023fine}
Yiwei Lyu, Tiange Luo, Jiacheng Shi, Todd~C Hollon, and Honglak Lee. 2023.
\newblock Fine-grained text style transfer with diffusion-based language models.
\newblock \emph{arXiv preprint arXiv:2305.19512}.

\bibitem[{Meng et~al.(2024)Meng, Xia, and Chen}]{meng2024simposimplepreferenceoptimization}
Yu~Meng, Mengzhou Xia, and Danqi Chen. 2024.
\newblock \href {https://arxiv.org/abs/2405.14734} {Simpo: Simple preference optimization with a reference-free reward}.
\newblock \emph{Preprint}, arXiv:2405.14734.

\bibitem[{Mudgal et~al.(2023)Mudgal, Lee, Ganapathy, Li, Wang, Huang, Chen, Cheng, Collins, Strohman et~al.}]{mudgal2023controlled}
Sidharth Mudgal, Jong Lee, Harish Ganapathy, YaGuang Li, Tao Wang, Yanping Huang, Zhifeng Chen, Heng-Tze Cheng, Michael Collins, Trevor Strohman, et~al. 2023.
\newblock Controlled decoding from language models.
\newblock \emph{arXiv preprint arXiv:2310.17022}.

\bibitem[{Nakano et~al.(2021)Nakano, Hilton, Balaji, Wu, Ouyang, Kim, Hesse, Jain, Kosaraju, Saunders et~al.}]{nakano2021webgpt}
Reiichiro Nakano, Jacob Hilton, Suchir Balaji, Jeff Wu, Long Ouyang, Christina Kim, Christopher Hesse, Shantanu Jain, Vineet Kosaraju, William Saunders, et~al. 2021.
\newblock Webgpt: Browser-assisted question-answering with human feedback.
\newblock \emph{arXiv preprint arXiv:2112.09332}.

\bibitem[{Ouyang et~al.(2022)Ouyang, Wu, Jiang, Almeida, Wainwright, Mishkin, Zhang, Agarwal, Slama, Ray et~al.}]{ouyang2022training}
Long Ouyang, Jeffrey Wu, Xu~Jiang, Diogo Almeida, Carroll Wainwright, Pamela Mishkin, Chong Zhang, Sandhini Agarwal, Katarina Slama, Alex Ray, et~al. 2022.
\newblock Training language models to follow instructions with human feedback.
\newblock \emph{Advances in neural information processing systems}, 35:27730--27744.

\bibitem[{Peng et~al.(2019)Peng, Kumar, Zhang, and Levine}]{peng2019advantage}
Xue~Bin Peng, Aviral Kumar, Grace Zhang, and Sergey Levine. 2019.
\newblock Advantage-weighted regression: Simple and scalable off-policy reinforcement learning.
\newblock \emph{arXiv preprint arXiv:1910.00177}.

\bibitem[{Rafailov et~al.(2024)Rafailov, Sharma, Mitchell, Manning, Ermon, and Finn}]{rafailov2024direct}
Rafael Rafailov, Archit Sharma, Eric Mitchell, Christopher~D Manning, Stefano Ermon, and Chelsea Finn. 2024.
\newblock Direct preference optimization: Your language model is secretly a reward model.
\newblock \emph{Advances in Neural Information Processing Systems}, 36.

\bibitem[{Santilli et~al.(2023)Santilli, Severino, Postolache, Maiorca, Mancusi, Marin, and Rodol{\`a}}]{santilli2023accelerating}
Andrea Santilli, Silvio Severino, Emilian Postolache, Valentino Maiorca, Michele Mancusi, Riccardo Marin, and Emanuele Rodol{\`a}. 2023.
\newblock Accelerating transformer inference for translation via parallel decoding.
\newblock \emph{arXiv preprint arXiv:2305.10427}.

\bibitem[{Schulman et~al.(2017)Schulman, Wolski, Dhariwal, Radford, and Klimov}]{schulman2017proximal}
John Schulman, Filip Wolski, Prafulla Dhariwal, Alec Radford, and Oleg Klimov. 2017.
\newblock Proximal policy optimization algorithms.
\newblock \emph{arXiv preprint arXiv:1707.06347}.

\bibitem[{Shen et~al.(2023)Shen, Jin, Huang, Liu, Dong, Guo, Wu, Liu, and Xiong}]{shen2023large}
Tianhao Shen, Renren Jin, Yufei Huang, Chuang Liu, Weilong Dong, Zishan Guo, Xinwei Wu, Yan Liu, and Deyi Xiong. 2023.
\newblock Large language model alignment: A survey.
\newblock \emph{arXiv preprint arXiv:2309.15025}.

\bibitem[{Song et~al.(2020)Song, Meng, and Ermon}]{song2020denoising}
Jiaming Song, Chenlin Meng, and Stefano Ermon. 2020.
\newblock Denoising diffusion implicit models.
\newblock \emph{arXiv preprint arXiv:2010.02502}.

\bibitem[{Song et~al.(2023)Song, Dhariwal, Chen, and Sutskever}]{song2023consistency}
Yang Song, Prafulla Dhariwal, Mark Chen, and Ilya Sutskever. 2023.
\newblock Consistency models.
\newblock \emph{arXiv preprint arXiv:2303.01469}.

\bibitem[{Stiennon et~al.(2020)Stiennon, Ouyang, Wu, Ziegler, Lowe, Voss, Radford, Amodei, and Christiano}]{stiennon2020learning}
Nisan Stiennon, Long Ouyang, Jeffrey Wu, Daniel Ziegler, Ryan Lowe, Chelsea Voss, Alec Radford, Dario Amodei, and Paul~F Christiano. 2020.
\newblock Learning to summarize with human feedback.
\newblock \emph{Advances in Neural Information Processing Systems}, 33:3008--3021.

\bibitem[{Team(2024)}]{qwen2.5}
Qwen Team. 2024.
\newblock \href {https://qwenlm.github.io/blog/qwen2.5/} {Qwen2.5: A party of foundation models}.

\bibitem[{Tunstall et~al.(2023)Tunstall, Beeching, Lambert, Rajani, Rasul, Belkada, Huang, von Werra, Fourrier, Habib et~al.}]{tunstall2023zephyr}
Lewis Tunstall, Edward Beeching, Nathan Lambert, Nazneen Rajani, Kashif Rasul, Younes Belkada, Shengyi Huang, Leandro von Werra, Cl{\'e}mentine Fourrier, Nathan Habib, et~al. 2023.
\newblock Zephyr: Direct distillation of lm alignment.
\newblock \emph{arXiv preprint arXiv:2310.16944}.

\bibitem[{Wang et~al.(2024)Wang, Xiong, Xie, Zhao, and Zhang}]{ArmoRM}
Haoxiang Wang, Wei Xiong, Tengyang Xie, Han Zhao, and Tong Zhang. 2024.
\newblock Interpretable preferences via multi-objective reward modeling and mixture-of-experts.
\newblock In \emph{EMNLP}.

\bibitem[{Wang et~al.(2023)Wang, Zhong, Li, Mi, Zeng, Huang, Shang, Jiang, and Liu}]{wang2023aligning}
Yufei Wang, Wanjun Zhong, Liangyou Li, Fei Mi, Xingshan Zeng, Wenyong Huang, Lifeng Shang, Xin Jiang, and Qun Liu. 2023.
\newblock Aligning large language models with human: A survey.
\newblock \emph{arXiv preprint arXiv:2307.12966}.

\bibitem[{Yang et~al.(2024{\natexlab{a}})Yang, Liu, Xie, Huang, Zhang, and Ananiadou}]{yang2024metaaligner}
Kailai Yang, Zhiwei Liu, Qianqian Xie, Jimin Huang, Tianlin Zhang, and Sophia Ananiadou. 2024{\natexlab{a}}.
\newblock Metaaligner: Towards generalizable multi-objective alignment of language models.
\newblock In \emph{The Thirty-eighth Annual Conference on Neural Information Processing Systems}.

\bibitem[{Yang et~al.(2024{\natexlab{b}})Yang, Zhang, Xia, Feng, Xiong, and Zhou}]{yang2024preference}
Shentao Yang, Shujian Zhang, Congying Xia, Yihao Feng, Caiming Xiong, and Mingyuan Zhou. 2024{\natexlab{b}}.
\newblock Preference-grounded token-level guidance for language model fine-tuning.
\newblock \emph{Advances in Neural Information Processing Systems}, 36.

\bibitem[{Yao et~al.(2023)Yao, Yi, Wang, Wang, and Xie}]{yao2023instructions}
Jing Yao, Xiaoyuan Yi, Xiting Wang, Jindong Wang, and Xing Xie. 2023.
\newblock From instructions to intrinsic human values--a survey of alignment goals for big models.
\newblock \emph{arXiv preprint arXiv:2308.12014}.

\bibitem[{Ye et~al.(2024{\natexlab{a}})Ye, Gao, Gong, Zheng, Jiang, Li, and Kong}]{ye2024beyond}
Jiacheng Ye, Jiahui Gao, Shansan Gong, Lin Zheng, Xin Jiang, Zhenguo Li, and Lingpeng Kong. 2024{\natexlab{a}}.
\newblock Beyond autoregression: Discrete diffusion for complex reasoning and planning.
\newblock \emph{arXiv preprint arXiv:2410.14157}.

\bibitem[{Ye et~al.(2024{\natexlab{b}})Ye, Gong, Chen, Zheng, Gao, Shi, Wu, Jiang, Li, Bi et~al.}]{ye2024diffusion}
Jiacheng Ye, Shansan Gong, Liheng Chen, Lin Zheng, Jiahui Gao, Han Shi, Chuan Wu, Xin Jiang, Zhenguo Li, Wei Bi, et~al. 2024{\natexlab{b}}.
\newblock Diffusion of thoughts: Chain-of-thought reasoning in diffusion language models.
\newblock \emph{arXiv preprint arXiv:2402.07754}.

\bibitem[{Zeng et~al.(2024)Zeng, Liu, Ma, Yang, Zhang, and Wang}]{zeng2024token}
Yongcheng Zeng, Guoqing Liu, Weiyu Ma, Ning Yang, Haifeng Zhang, and Jun Wang. 2024.
\newblock Token-level direct preference optimization.
\newblock \emph{arXiv preprint arXiv:2404.11999}.

\bibitem[{Zhang et~al.(2024)Zhang, Gu, Wu, Zhai, Susskind, and Jaitly}]{zhang2024planner}
Yizhe Zhang, Jiatao Gu, Zhuofeng Wu, Shuangfei Zhai, Joshua Susskind, and Navdeep Jaitly. 2024.
\newblock Planner: generating diversified paragraph via latent language diffusion model.
\newblock \emph{Advances in Neural Information Processing Systems}, 36.

\bibitem[{Zheng et~al.(2023)Zheng, Chiang, Sheng, Zhuang, Wu, Zhuang, Lin, Li, Li, Xing et~al.}]{zheng2023judging}
Lianmin Zheng, Wei-Lin Chiang, Ying Sheng, Siyuan Zhuang, Zhanghao Wu, Yonghao Zhuang, Zi~Lin, Zhuohan Li, Dacheng Li, Eric Xing, et~al. 2023.
\newblock Judging llm-as-a-judge with mt-bench and chatbot arena.
\newblock \emph{Advances in Neural Information Processing Systems}, 36:46595--46623.

\bibitem[{Zhong et~al.(2024)Zhong, Feng, Xiong, Cheng, Zhao, He, Bian, and Wang}]{zhong2024dpo}
Han Zhong, Guhao Feng, Wei Xiong, Xinle Cheng, Li~Zhao, Di~He, Jiang Bian, and Liwei Wang. 2024.
\newblock Dpo meets ppo: Reinforced token optimization for rlhf.
\newblock \emph{arXiv preprint arXiv:2404.18922}.

\end{thebibliography}

\newpage
\appendix
\section{Related Works}
\label{appendix: related works}
\subsection{Align LLM with Human Preference.}

A prominent approach to learning from human preferences is RLHF \cite{ouyang2022training, stiennon2020learning, christiano2017deep, bai2022training}. In this framework, a reward model is first trained, followed by the training of a bandit policy using Proximal Policy Optimization (PPO) \cite{schulman2017proximal}. 
Recent advancements such as direct preference optimization (DPO) \cite{rafailov2024direct, meng2024simposimplepreferenceoptimization, ethayarajh2024kto} optimize the bandit policy directly from human preferences, bypassing the need for a reward model. These approaches are simpler to implement and require fewer computational resources.
Inference-time approaches, on the other hand, achieve alignment by customizing the output of large language models (LLMs) during the decoding phase, without the need for parameter optimization. This results in enhanced flexibility and efficiency~\cite{khanov2024argsalignmentrewardguidedsearch, mudgal2023controlled, chen2024pad}. One representative method treats the text-generation process as a search problem, guided by external rewards \cite{huang2024deal, han2024value, chakraborty2024transfer}. Another category of methods focuses on learning to refine the generated text~\cite{li2023rain, ji2024aligner, yang2024metaaligner}.

\paragraph{Token and Sentence-level.}
Existing training-based or inference-time alignment approaches typically rely on token-level rewards, while human preferences are generally provided and defined at the sentence level \cite{li2023remax, ahmadian2024back, zeng2024token}. To address this discrepancy, some works \cite{lightman2023let, yang2024preference, zeng2024token} leverage token-wise or step-wise information to improve alignment performance. 
In contrast, this paper proposes modeling alignment as a sentence-level denoising process. We introduce a model-agnostic, inference-time alignment method, and our empirical results demonstrate its superiority in both performance and efficiency.

\subsection{Parallel Decoding and Diffusion Process.}
\paragraph{Parallel Decoding of LLMs}

Parallel decoding has been increasingly utilized and developed in recent research to accelerate the inference processes of large language models (LLMs). 
One line of research, including works by \citet{leviathan2023fast,chen2023accelerating}, focuses on speculative decoding. These techniques enhance LLM decoding speed by employing a smaller draft model to predict the outputs, which are then verified in parallel by a larger target model.
Another research trajectory explores parallel decoding strategies that do not rely on a draft model. Methods such as conditioning on ``look-ahead'' tokens or employing Jacobi iterations have been investigated by \citet{santilli2023accelerating, fu2024break}. These approaches allow the target model to produce several tokens simultaneously, aiming for rapid convergence to a fixed point on a Jacobi trajectory. CLLMs~\cite{song2023consistency} develop a novel approach, refining the target LLM to consistently predict the fixed point from any given state.

\begin{table*}[t]
\caption{\textbf{Comparison results of \model~models.} The experiments are conducted on base models of Qwen-2.5-7B and 14B. It shows that \model~consistently achieves superior performance across various base models.}
\label{appendix_tab:various_base_models}
\centering
\resizebox{.98\textwidth}{!}{
  \begin{tabular}{@{}l
  >{\centering\arraybackslash}m{1.3cm}
  >{\centering\arraybackslash}m{1.3cm}
  >{\centering\arraybackslash}m{1.3cm}
  >{\centering\arraybackslash}m{1.3cm}
  >{\centering\arraybackslash}m{1.3cm}
  >{\centering\arraybackslash}m{1.3cm}
  >{\centering\arraybackslash}m{1.3cm}
  >{\centering\arraybackslash}m{1.7cm}
  @{}}
    \toprule
     \multirow{2}{*}{\textbf{Base Models}} 
    &\multicolumn{3}{c}{\textbf{MT-bench}} & \multicolumn{2}{c}{\textbf{AlpacaEval 2}} &\multicolumn{3}{c}{\textbf{HH-RLHF}} \\
    \cmidrule(r){2-4} \cmidrule(lr){5-6} \cmidrule(lr){7-9} 
           & \textbf{1-Turn} & \textbf{2-Turn} & \textbf{Avg.} & \textbf{LC (\%)} & \textbf{WR~(\%)} & \textbf{Overall} & \textbf{Helpful} & \textbf{Harmless}   \\

\midrule
Qwen2.5-7B-Instruct & 7.11 & 6.96 & 7.03 & 45.03 & 49.95 & 0.1095 & 0.6995 & 0.9442 \\
\textit{w.}~DPO & 7.37 & 6.96 & 7.17 & 50.55 & 55.00 & 0.1109 & 0.7061 & 0.9460 \\
\textit{w.}~SimPO & 7.41 & 6.98 & 7.20 & 48.76 & 52.75 & 0.1100 & 0.7047 & 0.9387 \\
\textit{w.}~BPO & 6.90 & 6.16 & 6.53 & 29.44 & 39.85 & 0.1086 & 0.6765 & 0.9418 \\
\textit{w.}~BoN & 7.50 & 7.10 & 7.30 & 50.42 & 55.43 & 0.1159 & 0.7066 & 0.9440 \\
\textit{w.}~Aligner & 6.24 & 3.76 & 5.00 & 42.15 & 45.82 & 0.1088 & 0.6993 & 0.9438 \\
\textit{w.}~MetaAligner & 6.41 & 5.13 & 5.77 & 36.58 & 38.45 & 0.0995 & 0.6966 & 0.9422 \\
\midrule
\textit{w.}~\model-2B & 7.01 & 6.34 & 6.67 & 43.89 & 49.07 & 0.1074 & 0.7013 & 0.9659 \\
\textit{w.}~\model-9B & 7.62 & 7.10 & 7.35 & 57.89 & 63.01 & 0.1117 & 0.7100 & 0.9445 \\
\midrule
Qwen2.5-14B-Instruct & 7.24 & 6.71 & 6.98 & 51.60 & 57.10 & 0.1117 & 0.7100 & 0.9445 \\
\textit{w.}~BPO & 7.21 & 6.82 & 7.02 & 37.02 & 47.51 & 0.1005 & 0.6853 & 0.9323 \\
\textit{w.}~BoN & 7.49 & 6.98 & 7.24 & 54.92 & 59.02 & 0.1182 & 0.7163 & 0.9485 \\
\textit{w.}~Aligner & 6.14 & 4.11 & 5.13 & 45.14 & 47.24 & 0.1114 & 0.7099 & 0.9438 \\
\textit{w.}~MetaAligner & 6.24 & 5.73 & 5.99 & 41.25 & 43.23 & 0.1092 & 0.7074 & 0.9375 \\
\midrule
\textit{w.}~\model-2B & 7.08 & 6.33 & 6.71 & 43.70 & 48.76 & 0.1078 & 0.7017 & 0.9704 \\
\textit{w.}~\model-9B & 7.62 & 7.35 & 7.48 & 55.13 & 60.65 & 0.1136 & 0.7185 & 0.9759 \\
\arrayrulecolor{black}
\bottomrule
  \end{tabular}}
\end{table*}

\paragraph{Text Diffusion Models}
Diffusion models have demonstrated significant diversity and controllability in image generation~\cite{ho2020denoising, song2020denoising, dhariwal2021diffusion}. Recently, these models have been extended to text generation, as evidenced by the works of \cite{li2022diffusion, gong2022diffuseq, lovelace2024latent}.
In essence, diffusion models execute a multi-step denoising process that progressively transforms random noise into a coherent data sample. In the context of text, diffusion models can be considered an evolution of traditional iterative Non-Autoregressive models, as described by Gong et al. (2022). These models have demonstrated the ability to match or surpass Autoregressive (AR) models in terms of text perplexity~\cite{han2022ssd, gulrajani2024likelihood}, diversity~\cite{gong2023diffuseq, zhang2024planner}, and various sequence-to-sequence tasks~\cite{ye2024diffusion, ye2024beyond}.

\paragraph{Connection with \model}
In this paper, we are motivated by the goal of aligning Large Language Models (LLMs) with human values or intentions, as outlined in~\cite{yao2023instructions}. We define preferences at the sentence-level, focusing on the style or format of complete answers generated by the LLMs.
If we consider each iteration of parallel decoding as a transition between states, this bears a formal resemblance to discrete diffusion models. In \model, we leverage parallel decoding to implement sentence-level denoising, thereby enhancing the modeling of the alignment process.

The development of \model~is also inspired by Consistency Models~\cite{song2023consistency} and CLLMs~\cite{kou2024cllms}. Consistency models address the limitation of the slow iterative sampling process by mapping any point along the probability flow ODE of the diffusion process back to the original point in a single step. CLLMs propose accelerating LLM inference by mapping the intermediate process of LLM parallel decoding to the final process. Similar to these works, we optimize \model~with consistency loss, thus enabling model-agnostic alignment.

\section{Experiment}
\subsection{Experimental Setups}
\label{sec: Appendix Experimental Setups}

\paragraph{Training Details.}
As for the training set, we collect 6 generations from 6 base models (i.e., Llama-3-8B-Instruct, Llama-3-8B-SFT, Mistral-7B-SFT, Mistral-7B-Instruct, Gemma-2-2B-Instruct, Gemma-2-9B-Instruct). We then employ \textit{ArmoRM}~\cite{ArmoRM} to score these responses. The response with the highest score is selected as $\by^{(T)}$. The remaining five responses are ranked according to their scores to serve as $\by^{(0:T-1)}$. In the training process, at each iteration, we randomly sample $\by^t$ from $\by^{(0:T-1)}$ for optimization. We train \model~models using the following hyperparameters: a learning rate of 1e-9, a batch size of 1 and gradient accumulation steps of 4, a max sequence length of 1024, and a cosine learning rate schedule with 3\% warmup steps for 1 epoch. All the models are trained with an Adam optimizer. All the training experiments in this paper were conducted on 8×A100 GPUs.

\paragraph{Evaluation Details.}
For the MT-bench, we use GPT-4 as the judge model, following the default settings. The scores are based on a single-answer rating scale from 1 to 10. For AlpacaEval, we use GPT-4 Turbo as the judge model, which performs pairwise comparison of responses generated by GPT-4, each with the same maximum length. For HH-RLHF, we use ArmoRM for single-answer rating and report the overall score, along with the ``helpful" and ``harmless" scores, which are provided in dimensions 9 and 10, respectively.

\paragraph{Baseline Details.}
Implementation details for different baselines are as follows:
\begin{itemize}
\setlength{\itemsep}{2pt}
  \item MetaAligner: we use the open-sourced MetaAligner-7B model~\url{https://huggingface.co/MetaAligner/MetaAligner-HH-RLHF-7B} on Huggingface and follow its official inference guideline on Huggingface.
  \item DPO, SimPO: we directly use open-sourced models~\url{https://huggingface.co/princeton-nlp} on Huggingface, which are fine-tuned according to the recipes in SimPO~\cite{meng2024simposimplepreferenceoptimization}.
  \item Args: We reproduce Args according to \url{https://github.com/deeplearning-wisc/args/tree/main} by replacing the reward model with ArmoRM~\cite{ArmoRM}.
  \item Aligner: we use the open-sourced Aligner-7B~\url{https://huggingface.co/aligner/aligner-7b-v1.0} on Huggingface and follow its guideline on Huggingface.
  \item BPO: we use the open-sourced BPO model~\url{https://huggingface.co/THUDM/BPO} on Huggingface and follow its official inference on Huggingface.
\end{itemize}

\begin{figure*}[h] 
    \centering
    \includegraphics[width=\textwidth]{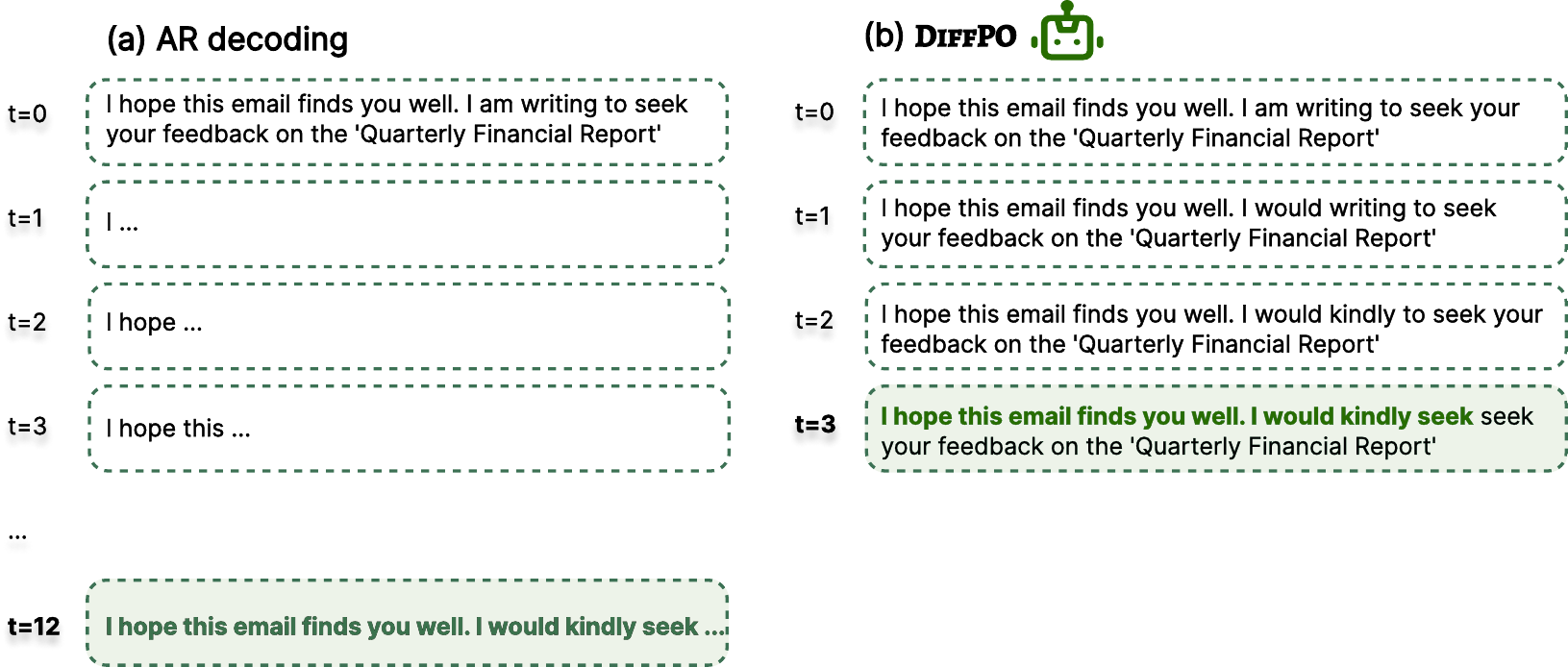}
    \caption{\textbf{Illustration of the Speedup of \model}. }
    \label{fig: parallel_decoding}
\end{figure*}

\subsection{Experimental Results}

\paragraph{\model~significantly outperforms existing preference optimization methods.}
We provided additional comparison with baselines, with results presented in Table~\ref{appendix_tab:various_base_models}. The experiments are conducted on base models of Qwen-2.5-7B and 14B.
While all preference optimization algorithms improve performance over the base model, \model~achieves the best overall performance across all benchmarks and settings. These consistent and significant improvements underscore the robustness and effectiveness of \model. Notably, \model~outperforms the training-based baselines (i.e., SimPO and DPO) across various settings, despite requiring only a single training session of \model~model and being capable of enhancing the performance of multiple base models.

\section{Analysis}

\subsection{Illustration of the Speed-up of \model}
As shown in Figure~\ref{fig: parallel_decoding}, AR decoding (e.g., Aligner~\cite{ji2024aligner}) typically generates only one aligned token per iteration. In contrast, \model~enables the skipping of satisfied tokens, thereby avoiding the time latency associated with token-level generation. As a result, \model can predict the modified subsequence in 3 iterations, achieving the same result as 11 iterations of AR decoding.

\section{More Related Works}
\label{sec:supp_survey}

\paragraph{LLM Pluralism and Fairness}
LLM alignment ensures AI systems follow human intentions and values~\cite{stiennon2020learning, bai2022training, ouyang2022training, achiam2023gpt}. However, within a single task, users' goals and values often differ. As AI systems are increasingly used by diverse groups, they must address a broader range of needs. In short, we need AI systems that are pluralistic and fair, being capable of reflecting diverse human values~\cite{chen2024fast, chen2024editable, fan2024biasalert, luo2024faintbench, fan2024fairmt, chen2024learnable}.



\end{document}